\documentclass[10pt,twocolumn,letterpaper]{article}

\usepackage{wacv}
\usepackage{times}
\usepackage{epsfig}
\usepackage{graphicx}
\usepackage{amsmath}
\usepackage{amssymb}
\usepackage[accsupp]{axessibility}  % Improves PDF readability for those with disabilities.

\usepackage{here}
\usepackage{caption}
\usepackage{multirow}
\usepackage[table,xcdraw]{xcolor}
\usepackage{arydshln}

\usepackage{graphics}
\usepackage{chngpage}
\definecolor{light-gray}{gray}{0.95}

\def\wacvPaperID{1137} % Enter the WACV Paper ID here

\wacvfinalcopy % *** Uncomment this line for the final submission

\ifwacvfinal
\fi

\ifwacvfinal
\usepackage[breaklinks=true,bookmarks=false]{hyperref}
\else
\usepackage[pagebackref=true,breaklinks=true,colorlinks,bookmarks=false]{hyperref}
\fi

\begin{document}

%%%%%%%%% TITLE
\title{Self-Supervised Learning of Domain Invariant Features\\ for Depth Estimation}

\author{%
  Hiroyasu Akada$^{1, 2}$ \,\,\,\,\, Shariq Farooq Bhat$^{1}$ \,\,\,\,\, Ibraheem Alhashim$^{3}$ \,\,\,\,\, Peter Wonka$^{1}$ \\
  \\
  $^{1}$KAUST, \,\,\,\, $^{2}$Keio University \,\,\,\,
  {\begin{tabular}[c]{@{}c@{}}$^{3}$National Center for Artificial Intelligence (NCAI),\\ Saudi Data and Artificial Intelligence Authority (SDAIA)\end{tabular}} \\
  {\tt\small hiroyasu5959@keio.jp} \,\,\,\, {\tt\small shariq.bhat@kaust.edu.sa} \,\,\,\, {\tt\small \{ibraheem.alhashim, pwonka\}@gmail.com}

% \and
}

\maketitle

\ifwacvfinal
\thispagestyle{empty}
\fi

%%%%%%%%% ABSTRACT
\begin{abstract}
  We tackle the problem of unsupervised synthetic-to-real domain adaptation for single image depth estimation. An essential building block of single image depth estimation is an encoder-decoder task network that takes RGB images as input and produces depth maps as output.
  In this paper, we propose a novel training strategy to force the task network to learn domain invariant representations in a self-supervised manner.
  Specifically, we extend self-supervised learning from traditional representation learning, which works on images from a single domain, to domain invariant representation learning, which works on images from two different domains by utilizing an image-to-image translation network.
  Firstly, we use an image-to-image translation network to transfer domain-specific styles between synthetic and real domains. This style transfer operation allows us to obtain similar images from the different domains. Secondly, we jointly train our task network and Siamese network with the same images from the different domains to obtain domain invariance for the task network. 
  Finally, we fine-tune the task network using labeled synthetic and unlabeled real-world data. Our training strategy yields improved generalization capability in the real-world domain. We carry out an extensive evaluation on two popular datasets for depth estimation, KITTI and Make3D. The results demonstrate that our proposed method outperforms the state-of-the-art on all metrics, \eg by 14.7\% on Sq Rel on KITTI. The source code and model weights will be made available.
\end{abstract}

%%%%%%%%% BODY TEXT
\section{Introduction}

\label{introduction}

Unsupervised domain adaptation (UDA) for single image depth estimation deals with the following problem: given a corpus of synthetic data (RGB images) and their labels (depth maps) together with real data (RGB images) without labels, the goal is to train a \emph{task network} (depth estimation network) to learn from the synthetic data in such a way so that it generalizes to real data. However, the large domain gap between real and synthetic data poses a significant challenge.

%-------------------------------------------------------------------------

\begin{figure}[t]
  \centering
  \begin{tabular}{cc}
  \begin{minipage}[c]{0.00001\hsize}
    \caption*{\rotatebox{90}{{ \,\,\,\,\,\,\,\,\,\,\,\, Ours \,\,\,\,\,\,\,\, CrDoCo*~\cite{CrDoCo} \,\,\,\, T$2$Net~\cite{zheng2018t2net} \,\,\,\,\,\,\,\,\,\,\,\, input \,\,\,\,\,\,\,\,}}}
  \end{minipage} &
  \begin{minipage}[c]{0.8\hsize}
    \includegraphics[width=1.0\linewidth]{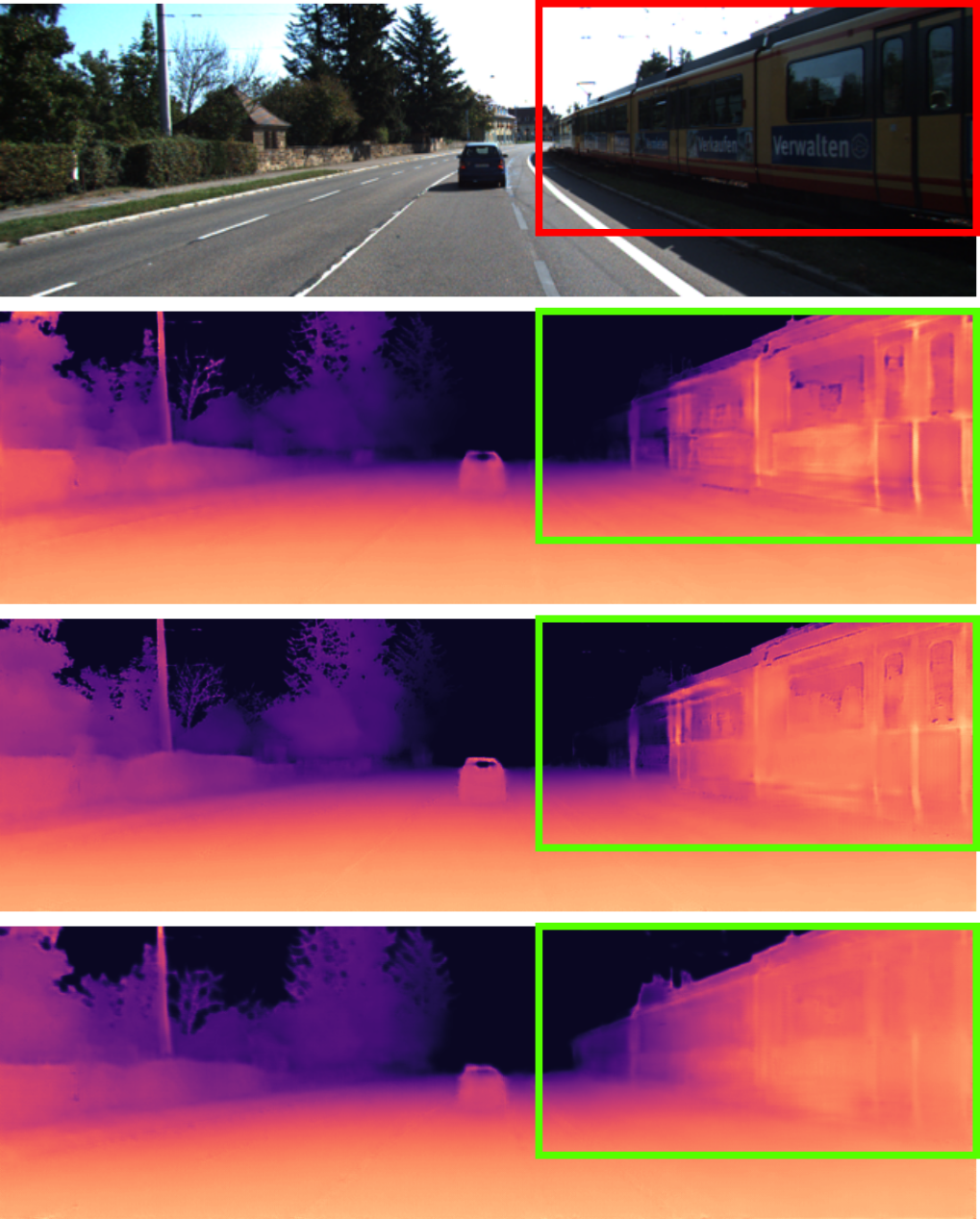}\\
  \end{minipage}
  \end{tabular}
  \vspace{-5pt}
  \caption{Predicted depth maps from our proposed method and comparison methods, T$2$Net~\cite{zheng2018t2net} and CrDoCo*~\cite{CrDoCo}. Our method is better at estimating consistent depth values on objects' surfaces than existing methods.}
  \label{fig:title}
\end{figure}

%-------------------------------------------------------------------------

We can identify three different strategies in UDA for single image depth estimation. The first strategy is to translate synthetic images to the real domain using an image-to-image translation (I2I) network and enforce a task network to learn representations only for the real domain~\cite{zheng2018t2net, lopez2020desc}. The second strategy also uses I2I for style transfer but enforces two separate task networks to learn separate representations for synthetic and real data respectively~\cite{CrDoCo, zhao2019geometry, pnvr2020sharingan}. 
In this strategy, the task network in the synthetic domain is trained in a supervised manner and is used to guide the other task network in the real domain. The third strategy is to train a task network to learn domain invariant features. The goal of such a strategy is to train a task network that can take both real and synthetic images as input and produce similar features for similar images from either domain. Our work follows this strategy as we try to train a single task network that can take input from both real and synthetic domains.

While the difference between these three strategies by itself is not that significant, we identified an important bottleneck that is common in all three of these strategies. The problem is that the task network needs to be initialized with reasonable (pre-trained) weights that work well in the real domain for the encoder, as indicated by~\cite{chen2021contrastive}, as well as the decoder. The initialization of the encoder can be easily done using the pre-trained weights on ImageNet~\cite{imagenet_cvpr09} that are widely available. However, initializing the decoder is challenging.
We found an elegant way to adapt recent works in self-supervised representation learning (SSRL)~\cite{chen2020exploring} to learn domain invariant features for the decoder. 
We propose several components that enable this adaptation. First, we replace data augmentation by an image-to-image translation network that learns mappings from synthetic-to-real and vice-versa. This lets us utilize similar images from either domain and use SSRL to enforce our task network to learn domain invariance. Second, we extend self-supervised learning to extract domain invariant representations for the decoder, coupled with the encoder pre-trained on ImageNet. 
Third, we use a channel-wise projector and predictor on high-resolution decoder features for self-supervision.

Our proposed framework consists of three stages: a style transfer stage, where we train the I2I networks, a self-supervised representation learning (SSRL) stage, which is used to learn a good initialization for the task network decoder, and a depth estimation stage, where the task network is fine-tuned using both labeled synthetic and unlabeled real-world datasets.

Extensive experiments show that our proposed method can lead to better generalization performance on the target domain and outperforms  state-of-the-art UDA methods for single image depth estimation on two popular datasets, KITTI~\cite{geiger2013vision} and Make3D~\cite{make3d} on all metrics.
In summary, we make the following main contributions:
\begin{itemize}
 \item We propose a novel UDA framework that enables an encoder-decoder monocular depth estimator to learn domain invariant representations and thus, generalizes well to the target domain.
 \item We devise a pre-training strategy for the decoder using self-supervised learning.
 \item We demonstrate that our proposed method achieves the new state-of-the-art in UDA for single image depth estimation on two popular datasets, KITTI~\cite{geiger2013vision} and Make3D~\cite{make3d}.
\end{itemize}

\section{Related Work}

\subsection{Supervised Depth Estimation}
Supervised learning methods are currently the top-performing approaches for the depth estimation task~\cite{NIPS2014_7bccfde7,Alhashim2018,gan2018monocular,fu2018deep,yin2019enforcing,lee2019big,bhat2020adabins}. 
%DenseDepth~\cite{Alhashim2018} leverages an ImageNet pre-trained encoder within their encoder-decoder network and tackles depth estimation as a regression problem.
BTS~\cite{lee2019big} proposes to utilize local planar guidance layers to effectively guide feature maps to full resolution instead of using conventional upsampling layers in their decoder blocks.
DORN~\cite{fu2018deep} considers depth estimation as a classification task by dividing the depth range into multiple bins that are fixed with predetermined widths. More recently, AdaBins~\cite{bhat2020adabins}, expands on DORN by introducing a transformer-based architecture to dynamically change the depth bins based on the input. While these supervised methods show promising results, they require fully labeled real-world datasets that are hard to prepare.

\subsection{UDA for Depth Estimation}
Unsupervised domain adaptation methods aim to learn from synthetic depth maps that are much easier to produce. The core idea of such methods is to align the data distribution between a synthetic dataset with full labels (\ie source domain) and a real-world dataset without labels (\ie target domain). In depth estimation, these works can be divided into two groups. The first group applies GAN~\cite{goodfellow2014generative}-based image translation techniques with extra information, such as real-world stereo pair images~\cite{zhao2019geometry, pnvr2020sharingan, DBLP:journals/corr/abs-1909-03943, DBLP:journals/corr/abs-1807-10915}, semantic segmentation images~\cite{lopez2020desc}, monocular video~\cite{Zhang_2020_CVPR}, pose data~\cite{kundu2018adadepth}, shading information~\cite{Bi_2019_ICCV}, or a small amount of real ground truth labels~\cite{zhao2020domain}. The second group, including our work, follows a fully unsupervised approach without such additional information. These methods also utilize adversarial training~\cite{goodfellow2014generative} to align data distributions at both feature and image levels~\cite{zheng2018t2net, CrDoCo}.

Our work is based on the state-of-the-art UDA method, CrDoCo~\cite{CrDoCo} and we introduce a novel training strategy that allows the depth estimator to learn domain invariance in a self-supervised manner.

%-------------------------------------------------------------------------

\begin{figure*}[t]
  \centering
  \includegraphics[width=1.0\linewidth]{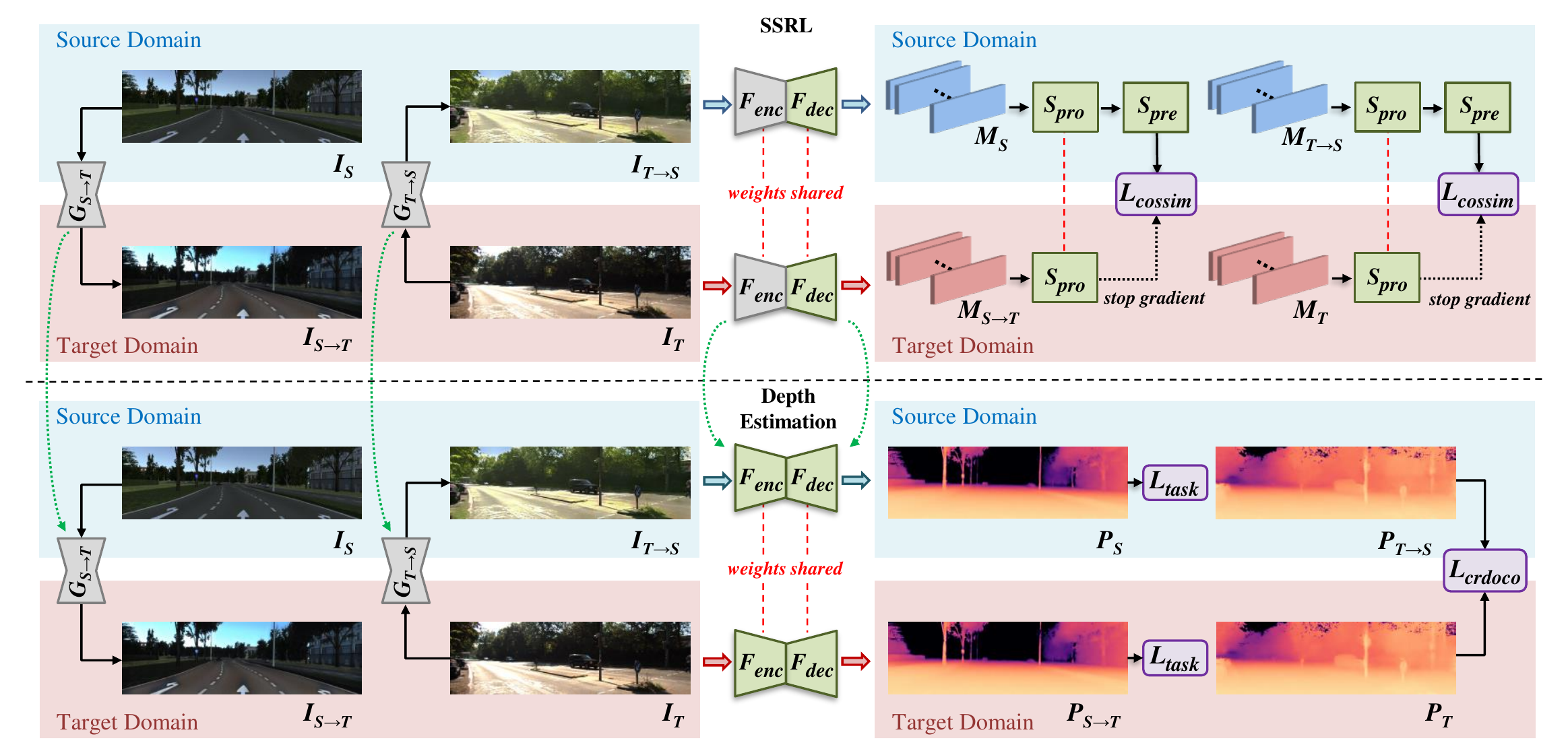}
  \caption{Overview of our proposed SSRL stage. Modules in green are trainable and modules in gray are non-trainable. Green dot arrows indicate that modules will be used in the next stage. As a preliminary step, we first train the image translation networks $G_{S\rightarrow T}$ and $G_{T\rightarrow S}$ in the style transfer stage as in most UDA methods including CrDoCo~\cite{CrDoCo}. The trained image translation network $G_{S\rightarrow T}$ and $G_{T\rightarrow S}$ are used in the SSRL and depth estimation stages with their weights fixed. As our main contribution, we introduce the self-supervised representation learning (SSRL) stage to enforce our task network to learn a domain invariant representation via our Siamese network in a self-supervised way. Note that $L_{cossim}$ is a symmetrized loss using corresponding feature maps (\eg $M_S$ and $M_{S\rightarrow T}$). Lastly, we fine-tune the final task network using the synthetic and real datasets in the depth estimation stage. Please see Section~\ref{preliminaries} and~\ref{SSRL} for more details.}
  \label{fig:overview}
\end{figure*}

%-------------------------------------------------------------------------

\subsection{Representation Learning}
Representation learning has been an actively researched domain in deep learning. One line of research in this area is contrastive learning. Contrastive learning methods introduce a contrastive loss~\cite{hadsell2006dimensionality} and a projection multi-layer perceptron (MLP) to make latent feature representations similar for similar input pairs (\ie positive pairs) and dissimilar for dissimilar input pairs (\ie negative pairs)~\cite{chen2020simple, NEURIPS2020_fcbc95cc, he2020momentum, chen2020improved, chen2021empirical}. Although contrastive learning have showed promising results, 
%it requires a large number of batch size that are difficult to set in dense prediction tasks. 
%
their training is prohibitively expensive within a UDA framework due to the use of a large batch size.   
Other representation learning methods, such as BYOL~\cite{NEURIPS2020_f3ada80d}, utilizes a momentum-based network coupled with a prediction MLP without the need of negative pairs. In addition, SimSiam~\cite{chen2020exploring} significantly simplifies the previously proposed networks and introduces only a prediction MLP and stop-gradient operation for representation learning. More recently, PixPro~\cite{xie2020propagate} is proposed to utilize pixel-level consistency on downstream dense prediction tasks, such as semantic segmentation.

We build our framework based on the prediction network and stop-gradient operation~\cite{chen2020exploring} to be trainable on images in one domain and the corresponding translated images in the other domain to obtain domain invariant representations.

\section{UDA for single image depth estimation}
\label{preliminaries}

We address the task of UDA for single image depth estimation. In terms of data availability, we have access to a set of synthetic RGB images and depth maps as source image data $I_S \in X_S$ and source label data $I_{S, lab} \in X_{S, lab}$, and a set of real-world RGB images as target image data $I_T \in X_T$. We aim at training a depth estimator without using real-world depth maps so that it is able to estimate an accurate depth map given a real-world single RGB image.

To this end, we propose a novel UDA framework that allows an encoder-decoder based depth estimator to learn a domain invariant representation via adversarial training and representation learning. 
% Our framework is based on three stages: a style transfer stage, a self-supervised representation learning (SSRL) stage, and a depth estimation stage as shown in Fig~\ref{fig:overview}.
Our main contribution lies in the introduction of the SSRL stage into current state-of-the-art UDA methods that are usually based on the style transfer stage and the depth estimation stage (task-specific stage) only.
We build on CrDoCo~\cite{CrDoCo} as the current state-of-the-art UDA method. In the following, we describe our method in more details.
First, we describe our modifications to CrDoCo to make it better compatible with our SSRL stage. Second, we provide details on our SSRL stage in Section~\ref{SSRL}.

%-------------------------------------------------------------------------

\subsection{Review of CrDoCo}
\label{sec:CrDoCo}

CrDoCo~\cite{CrDoCo} is composed of the style transfer and depth estimation stages.
In the style transfer stage, CrDoCo~\cite{CrDoCo} trains a bidirectional image-to-image translation network~\cite{CycleGAN2017} to learn a mapping between source and target domains via adversarial training.
In the depth estimation stage, the image translators are then used together with two feature discriminators and two task networks~\cite{zheng2018t2net} that are deployed in source and target domains. Specifically, the task network in each domain takes as inputs images with domain-specific styles (\eg the task network in the source domain takes as inputs source images and translated images from the target domain). Along with label supervision, CrDoCo~\cite{CrDoCo} introduces a cross-domain consistency loss $L_{crdoco}$ to enforce prediction level alignment for unlabeled images. Please see~\cite{CrDoCo} for more details on their architecture. 

\subsection{Modifications}
\label{modifications}

Following CrDoCo~\cite{CrDoCo}, we adopt the bidirectional image-to-image translation network~\cite{CycleGAN2017} in our framework, \ie two generators $G_{S\rightarrow T}$, $G_{T\rightarrow S}$, and two discriminators $D_{S}$, $D_{T}$. However, we remove the two feature discriminators for faster training and modify their task networks into a weights-shared model $F$, which is crucial to leverage our proposed SSRL stage described in Section~\ref{SSRL}. 

We also design our encoder-decoder based task network with EfficientNetB5~\cite{tan2019efficientnet} as an encoder instead of using older encoders~\cite{zheng2018t2net, CrDoCo, zhao2019geometry, pnvr2020sharingan}. 
In addition, we pre-train the encoder on ImageNet~\cite{imagenet_cvpr09}, which is a recent de facto standard for many UDA methods for depth estimation~\cite{Alhashim2018, gan2018monocular, fu2018deep, yin2019enforcing, lee2019big, bhat2020adabins} as well as semantic segmentation~\cite{liu2021domain, subhani2020learning, yang2020fda, li2020content, lv2020cross, lian2019constructing, chen2019domain, zou2018domain, wu2018dcan, sankaranarayanan2018learning, zhu2018penalizing}. 
% Pre-training of the encoder is a recent de facto standard for performance improvement in supervised methods for depth estimation~\cite{Alhashim2018, gan2018monocular, fu2018deep, yin2019enforcing, lee2019big, bhat2020adabins} and other UDA methods for semantic segmentation~\cite{liu2021domain, subhani2020learning, yang2020fda, li2020content, lv2020cross, lian2019constructing, chen2019domain, zou2018domain, wu2018dcan, sankaranarayanan2018learning, zhu2018penalizing}.
% The effectiveness of the ImageNet initialization in dense prediction tasks have already been shown in many previous works. For example, \cite{monodepth17, monodepth2} demonstrate that the pre-trained models outperform the baseline models by significant margins on all metrics in depth estimation. \cite{chen2021contrastive} also describes the importance of ImageNet initialization in UDA for image classification and semantic segmentation. Additionally, using ImageNet initialization enables us to see the exact performance gap between the current state-of-the-art UDA and supervised depth estimation methods. Even though all recent UDA approaches for depth estimation~\cite{zheng2018t2net, CrDoCo, zhao2019geometry, pnvr2020sharingan} have not utilized ImageNet initialization, because of the reasons above, we use the pre-trained encoder for all experiments including our comparative methods for a fair comparison. Please see Section~\ref{comparisons} for more detail.

For the style transfer and depth estimation stages, we adopt the same loss functions as CrDoCo~\cite{CrDoCo}. Please see \cite{CrDoCo} and our supplementary material for more details on each stage and on the loss functions.

%-------------------------------------------------------------------------

\section{Proposed SSRL stage}
\label{SSRL}

In this section, we describe our proposed SSRL stage as shown in Fig.~\ref{fig:overview}. In the SSRL stage, we utilize the image-to-image translation network $G_{S\rightarrow T}$ and $G_{T\rightarrow S}$, the encoder-decoder based image-to-depth task network $F$, and a Siamese network $S$. Note that the image-to-image translation networks $G_{S\rightarrow T}$ and $G_{T\rightarrow S}$ are pre-trained in the style transfer stage as in CrDoCo~\cite{CrDoCo}.

The purpose of the SSRL stage is to force $F$ to learn a latent feature representation that is invariant to synthetic (\ie source) and real (\ie target) domains so that $F$ can generalize well in the real domain. In the architecture of $F$, its encoder $F_{enc}$ is initialized by ImageNet~\cite{imagenet_cvpr09} pre-training and thus, it is already good at extracting features in the real domain for dense prediction tasks as indicated in~\cite{monodepth17, monodepth2, chen2021contrastive}. Therefore, we aim to train its decoder $F_{dec}$ jointly with $S$ so that $F_{dec}$ can also perform well in the real domain. Here, we fix the weights of $F_{enc}$. 
% In addition, we exploit $G_{S\rightarrow T}$ and $G_{T\rightarrow S}$ that are trained in the style transfer stage and fix the the weights of $G_{S\rightarrow T}$ and $G_{T\rightarrow S}$ for faster training.

%-------------------------------------------------------------------------

\begin{table*}[t]
\begin{center}
\scalebox{0.83}{
\begin{tabular}{lcccccccccc}
\hline
\multicolumn{1}{l||}{}                                      & \multicolumn{1}{c|}{}                          & \multicolumn{1}{c|}{}                                                                    & \multicolumn{1}{c|}{}                                                                                      & \multicolumn{4}{c|}{Lower is better}                                                                   & \multicolumn{3}{c}{Higher is better}                               \\ \cline{5-11} 
\multicolumn{1}{l||}{\multirow{-2}{*}{Method}}              & \multicolumn{1}{c|}{\multirow{-2}{*}{Train}}   & \multicolumn{1}{c|}{\multirow{-2}{*}{Dataset}}                                           & \multicolumn{1}{c|}{\multirow{-2}{*}{\begin{tabular}[c]{@{}c@{}}Evaluation \\ resolution\end{tabular}}}    & Abs Rel              & Sq Rel               & RMSE                 & \multicolumn{1}{c|}{RMSE log}     & $\delta < 1.25$      & $\delta < 1.25^2$    & $\delta < 1.25^3$    \\ \hline
\multicolumn{1}{l||}{BTS \cite{lee2019big}}                 & \multicolumn{1}{c|}{S}                         & \multicolumn{1}{c|}{K(I+D)}                                                              & \multicolumn{1}{c|}{$1241\times 376$}                                                                      & 0.059                & 0.245                & 2.756                & \multicolumn{1}{c|}{0.096}        & 0.956                & 0.993                & 0.998                \\
\multicolumn{1}{l||}{AdaBins \cite{bhat2020adabins}}        & \multicolumn{1}{c|}{S}                         & \multicolumn{1}{c|}{K(I+D)}                                                              & \multicolumn{1}{c|}{$1241\times 376$}                                                                      & 0.058                & 0.190                & 2.360                & \multicolumn{1}{c|}{0.088}        & 0.964                & 0.995                & 0.999                \\ \hline
\multicolumn{1}{l||}{Monodepth2 \cite{monodepth2}}          & \multicolumn{1}{c|}{SS}                        & \multicolumn{1}{c|}{\begin{tabular}[c]{@{}c@{}}K(I)\\ + video + stereo\end{tabular}}     & \multicolumn{1}{c|}{$1024\times 320$}                                                                      & 0.106                & 0.806                & 4.630                & \multicolumn{1}{c|}{0.193}        & 0.876                & 0.958                & 0.980                \\
\multicolumn{1}{l||}{Monodepth2 \cite{monodepth2}}          & \multicolumn{1}{c|}{SS}                        & \multicolumn{1}{c|}{\begin{tabular}[c]{@{}c@{}}K(I) + video\end{tabular}}                & \multicolumn{1}{c|}{$1024\times 320$}                                                                      & 0.115                & 0.882                & 4.701                & \multicolumn{1}{c|}{0.190}        & 0.879                & 0.961                & 0.982                \\
\multicolumn{1}{l||}{Johnston \cite{johnston2020self}}      & \multicolumn{1}{c|}{SS}                        & \multicolumn{1}{c|}{\begin{tabular}[c]{@{}c@{}}K(I) + video\end{tabular}}                & \multicolumn{1}{c|}{$640\times 192$}                                                                       & 0.106                & 0.861                & 4.699                & \multicolumn{1}{c|}{0.185}        & 0.889                & 0.962                & 0.982                \\ \hline
\rowcolor[HTML]{E8E3E3} 
\multicolumn{1}{l||}{T$^2$Net \cite{zheng2018t2net}}        & \multicolumn{1}{c|}{U(DA)}                     & \multicolumn{1}{c|}{vK(I+D)+K(I)}                                                        & \multicolumn{1}{c|}{$960\times 288$}                                                                       & 0.179                & 1.620                & 6.108                & \multicolumn{1}{c|}{0.257}        & 0.754                & 0.902                & 0.962                \\

\rowcolor[HTML]{E8E3E3} 
\multicolumn{1}{l||}{CrDoCo \cite{CrDoCo}}                  & \multicolumn{1}{c|}{U(DA)}                     & \multicolumn{1}{c|}{vK(I+D)+K(I)}                                                        & \multicolumn{1}{c|}{$960\times 288$}                                                                       & 0.232                & 2.204                & 6.733                & \multicolumn{1}{c|}{0.291}        & 0.739                & 0.883                & 0.942                \\
\rowcolor[HTML]{E8E3E3} 
\multicolumn{1}{l||}{CrDoCo* \cite{CrDoCo}}                  & \multicolumn{1}{c|}{U(DA)}                     & \multicolumn{1}{c|}{vK(I+D)+K(I)}                                                        & \multicolumn{1}{c|}{$960\times 288$}                                                                      & 0.174                & 1.439                & 5.701                & \multicolumn{1}{c|}{0.241}        & 0.770                & 0.914                & 0.967                \\

\rowcolor[HTML]{E8E3E3} 
\multicolumn{1}{l||}{Ours}                                  & \multicolumn{1}{c|}{U(DA)}                     & \multicolumn{1}{c|}{vK(I+D)+K(I)}                                                        & \multicolumn{1}{c|}{$960\times 288$}                                                                       & \bf 0.168            & \bf 1.228            & \bf 5.498            & \multicolumn{1}{c|}{\bf 0.235}    & \bf 0.771            & \bf 0.921            & \bf 0.973            \\ \hline
\end{tabular}
}
\end{center}
\caption{Quantitative results on KITTI~\cite{geiger2013vision}. Methods, which use only synthetic image-depth pairs and real depth maps as training datasets, are marked in gray. 
For training, S: supervised, SS: self-supervised, U: unsupervised, and DA: domain adaptation. For dataset, K and vK represent KITTI and virtual KITTI, I and D indicate the use of RGB images and depth maps, stereo and video indicate the use of stereo and video information. The model with $^{*}$ is trained with the same hyper-parameters as our model.}
\label{vkitti2kitti}
\end{table*}

%-------------------------------------------------------------------------

\begin{table}[t]
\begin{center}
\scalebox{0.75}{

\begin{tabular}{l||c|cccc}
\hline
                                                &                         & \multicolumn{4}{c}{Lower is better} \\ \cline{3-6} 
\multirow{-2}{*}{Method}                        & \multirow{-2}{*}{Train} & Abs Rel & Sq Rel & RMSE   & RMSE log \\ \hline
Zhou \cite{zhou2017unsupervised}                & SS                      & 0.383   & 5.321  & 10.470 & 0.478    \\
DDVO \cite{wang2018learning}                    & SS                      & 0.387   & 4.720  & 8.090  & 0.204    \\
Monodepth2 \cite{monodepth2}                    & SS                      & 0.322   & 3.589  & 7.417  & 0.163    \\
Johnston \cite{johnston2020self}                & SS                      & 0.297   & 2.902  & 7.013  & 0.158    \\ 
% FAL~\cite{NEURIPS2020_951124d4}                 & SS                      & 0.297   & 2.913  & 6.810  & -        \\ 
\hline
\rowcolor[HTML]{E8E3E3} 
T2Net~\cite{zheng2018t2net}                     & U(DA)                   & 0.337 & 4.767 & 8.735 & 0.128    \\
\rowcolor[HTML]{E8E3E3} 
CrDoCo~\cite{CrDoCo}                            & U(DA)                   & 0.606 & 14.221 & 13.92 & 0.209    \\
\rowcolor[HTML]{E8E3E3} 
CrDoCo*~\cite{CrDoCo}                           & U(DA)                   & 0.330 & 4.295 & 8.011 & 0.127    \\ 
\rowcolor[HTML]{E8E3E3} 
Ours                                            & U(DA)                   & \bf 0.309 & \bf 3.567 & \bf 7.401 & \bf 0.119   \\ \hline
\end{tabular}
}
\end{center}
\caption{Quantitative results on Make3D~\cite{make3d}.}
\label{make3d}
\end{table}

%-------------------------------------------------------------------------

\subsection{Components of the proposed SSRL stage}

The first adaptation from traditional work in SSRL is to replace data augmentation with style transfer. Instead of learning identical representations that are invariant to a given set of image transformations such as color shifts, we aim to learn domain invariant representations for synthetic and real images. We can use two types of image pairs as input: $I_{S\rightarrow T}$ together with $I_S$ and $I_{T\rightarrow S}$ together with $I_T$. 
In our notation, $I_{S\rightarrow T}$ and $I_{T\rightarrow S}$ are the outputs of translation network $G_{S\rightarrow T}$ and $G_{T\rightarrow S}$, respectively.

% with synthetic $I_S$ and real images $I_T$ as inputs to obtain corresponding translated images $I_{S\rightarrow T}$ and $I_{T\rightarrow S}$, respectively. These four types of images are forwarded into $F$.

Different from previously proposed representation learning methods~\cite{NEURIPS2020_f3ada80d, chen2020exploring, xie2020propagate}, we aim to learn representations in the decoder. This leaves two design choices to explore. First, we need to decide which representations in the decoder to target. The final output \ie the depth values are no longer useful as representations. The obvious choice would be the features in the last layer before the output layer, but earlier features could also yield better results. The decision of which layer to use for SSRL is taken using an empirical study described in Sec.~\ref{sec:ablation}. Second, different from previous work, our feature maps have a much higher resolution.
In our recommended architecture, we extract corresponding high dimensional feature maps before the final output layer (convolutional layer) from $F_{dec}$, \ie $M_S$, $M_T$, $M_{S\rightarrow T}$, and $M_{T\rightarrow S}$ as shown in Fig~\ref{fig:overview}. The size of each feature map is $C \times W \times H$, where $C = 12$, $W = 960$, $H = 288$ in our experiment. Here, we consider $M_S$ and $M_{S\rightarrow T}$ (or $M_T$ and $M_{T\rightarrow S}$ ) as a set of features of augmented views from the same scene. In the following parts, we only describe the domain invariant representation learning on $M_S$ and $M_{S\rightarrow T}$ for brevity. The second case, when using $M_T$ and $M_{T\rightarrow S}$, works in the same way. 

We also developed a component that corresponds to the projector in traditional SSRL~\cite{NEURIPS2020_f3ada80d, chen2020exploring, xie2020propagate}. In contrast to this traditional setting, we do not reduce the features dimension.
In our Siamese network $S$, the projector $S_{pro}$ takes the feature maps $M_S$ and $M_{S\rightarrow T}$ as inputs to output corresponding feature embeddings of the same size $C \times W \times H$.
Similar to the traditional SSRL architecture, the predictor $S_{pre}$ aims to transform the embedding of one view and matches it to the other view. We denote these outputs as  $z_{S} \triangleq S_{pro}(M_S)$, $p_{S} \triangleq S_{pre}(S_{pro}(M_S))$, $z_{S\rightarrow T} \triangleq S_{pro}(M_{S\rightarrow T})$, and $p_{S\rightarrow T} \triangleq S_{pre}(S_{pro}(M_{S\rightarrow T}))$. Here, the $S_{pro}$ and $S_{pre}$ share the weights for $M_S$ and $M_{S\rightarrow T}$ as well as $M_T$ and $M_{T\rightarrow S}$.
However, our predictor architecture incorporates a bottleneck. This follows previous architectures that also incorporate a dimension reduction component in the predictor.
Different from previous work, both our projector and predictor are per-pixel MLPs (convolutions with a $1 \times 1$ kernel).

\subsection{Loss function of the proposed SSRL stage}
\label{loss function in SSRL}

We follow the previous methods~\cite{NEURIPS2020_f3ada80d, chen2020exploring} by using a negative cosine similarity as our loss function in the SSRL stage. However, unlike the previous methods, the latent features from $F_{dec}$ are high dimensional and therefore, we assume that it is beneficial to take into consideration the correspondence between the features at the level of pixels. 
Based on this assumption, we define a symmetrized loss as 

\begin{equation}
\label{loss:cossim}
\begin{split}
 L_{cossim} &= \frac{1}{2}(cossim(p_{S}, sg(z_{S\rightarrow T}))) \\
             &+ \frac{1}{2}(cossim(p_{S\rightarrow T}, sg(z_{S}))),
\end{split} 
\end{equation}

where $sg$ is a stop gradient operation~\cite{chen2020exploring} and $cossim(a, b)$ is a pixel-wise negative cosine similarity function defined as

\begin{equation}
\label{operation:cossim}
\begin{split}
 cossim(A, B) &= \frac{1}{N} \sum_{i \in N}^{} \left(- \frac{a_{i}}{\| a_{i} \|_{2}} \cdot \frac{b_{i}}{\| b_{i} \|_{2}} \right),
\end{split} 
\end{equation}

where $a_{i}$ and $b_{i}$ are feature vectors of the size $C \times 1 \times 1$ with the spatial index $i$ of feature maps $A$ and $B$ (\ie $a_{i} \in A$ and $b_{i} \in B$), and $N$ is the total number of the spatial indices. 
% Please see Table~\ref{ablation study cossim} for our ablation studies on $L_{cossim}$.
%
Overall, we minimize $L_{cossim}$ in a self-supervised manner in the SSRL stage, \ie $L_{SSRL} = L_{cossim}$.

% \begin{equation}
% \label{loss:stage2}
% \begin{split}
%  L_{stage 2} &= L_{cossim}
% \end{split} 
% \end{equation}

%-------------------------------------------------------------------------

\begin{figure*}[t]
  \centering
  \begin{tabular}{cc}
  \begin{minipage}[c]{0.00001\hsize}
    \caption*{\rotatebox{90}{{\small \,\,\,\,\,\,\,\,\,\,\,\, Ours \,\,\,\, CrDoCo*~\cite{CrDoCo} \, T$2$Net~\cite{zheng2018t2net} \,\,\,\,\,\,\, input \,\,\,\,\,\,\,\,}}}
  \end{minipage} &
  \begin{minipage}[c]{0.95\hsize}
    \includegraphics[width=1.0\linewidth]{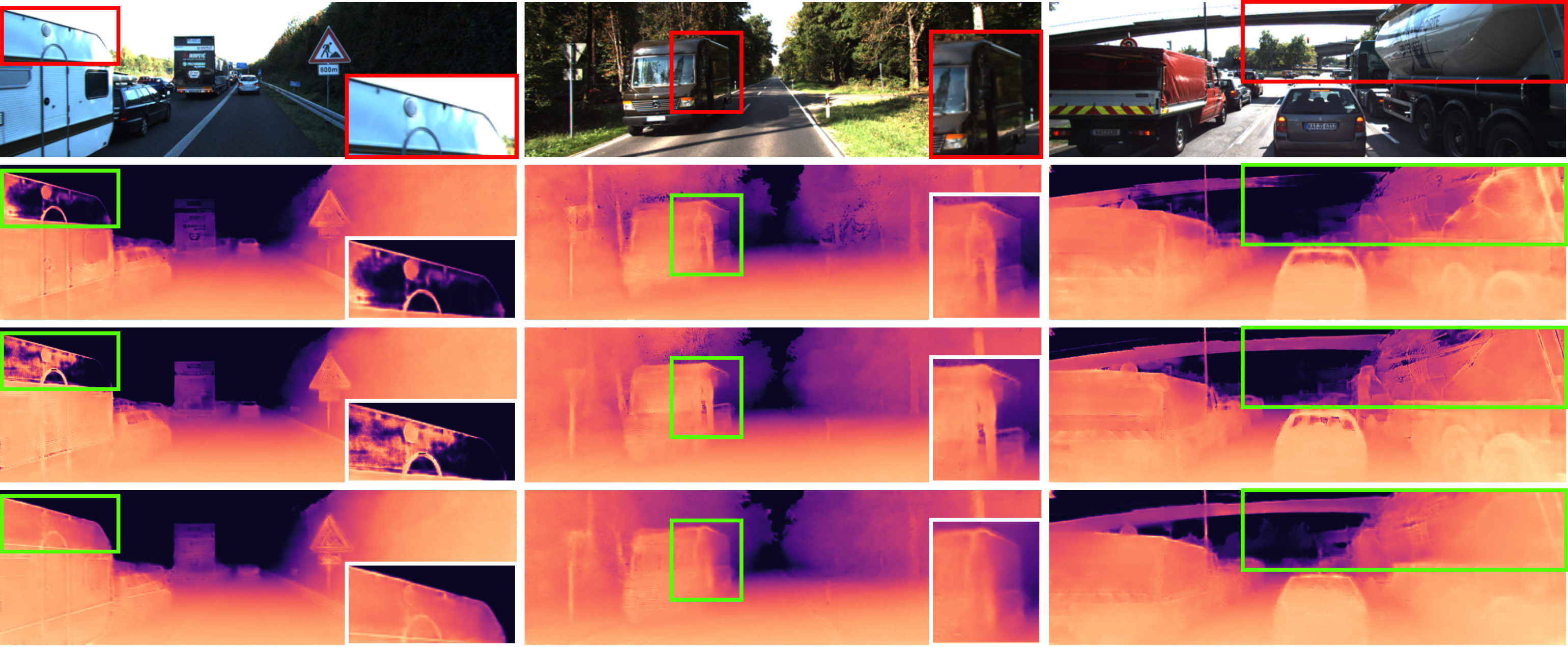}\\
  \end{minipage}
  \end{tabular}
  \vspace{-10pt}
  \caption{Qualitative results on KITTI~\cite{geiger2013vision}. Additional qualitative results are available in the supplementary material.}
  \label{fig:success}
\end{figure*}

%-------------------------------------------------------------------------

%-------------------------------------------------------------------------
\section{Experiments and results}

\label{experiments and results}

\subsection{Network architecture}
\label{network architecture}

We use CycleGAN~\cite{CycleGAN2017} as our bidirectional image translation network $G_{S\rightarrow T}$, $G_{T\rightarrow S}$, $D_{S}$, and $D_{T}$. For our task network $F$, we adopt an encoder-decoder based architecture using EffcientNet-B5~\cite{tan2019efficientnet} pre-trained on ImageNet~\cite{imagenet_cvpr09} as the encoder. For our Siamese network, we follow ~\cite{NEURIPS2020_f3ada80d, chen2020exploring} to leverage a projector and a predictor. Specifically, our projector $S_{pro}$ has 3 convolutional layers with $1\times 1$ kernels, ReLU activation and Batch Normalization~\cite{ioffe2015batch} after each layer except the last layer. Since the output feature maps from $F_{dec}$ in the SSRL stage are the size of $C (=12) \times W \times H$, we set the channel dimension of each layer in the projector to $C$. Also, the predictor has 2 convolutional layers with $1\times 1$ kernels, ReLU and Batch Normalization~\cite{ioffe2015batch} applied only after the first layer. In our predictor $S_{pre}$, the input and output channel dimensions are set to $C$ whereas the hidden layer's channel dimension is set to $C/\alpha$, where $\alpha$ is a scaling factor. Here, we set $\alpha = 3$ in our experiments. Please see Table~\ref{ablation study hidden layer} for our ablation studies on $\alpha$ and implementation code for more details.

\subsection{Datasets}

In our experiments, we use KITTI~\cite{geiger2013vision} as the target domain dataset and virtual KITTI (vKITTI)~\cite{gaidon2016virtual} as the source domain dataset. KITTI is an outdoor scene dataset captured using a moving vehicle with a resolution of around $1241 \times 376$. We use a subset of 22,600 images for training as specified by Eigen \etal~\cite{NIPS2014_7bccfde7}. vKITTI provides 21,260 synthetic image-depth pairs generated from different virtual urban worlds. The maximum sensed depth in KITTI is on the order of 80m while vKITTI has more precise depth values to a maximum of 655.3m. Consistent with previous work~\cite{zheng2018t2net}, we remove 'fog' and 'rain' images, and clip the maximum depth in vKITTI to match that of KITTI, \ie 80m.

For input resolution, we use relatively larger input size that is comparable with the state-of-the-art supervised methods (\eg $704 \times 352$ cropped from $1241 \times 376$ for training and $1241 \times 376$ for testing)~\cite{lee2019big, bhat2020adabins} to see the current performance gap between UDA methods and supervised counterparts. Specifically, images are resized to $960 \times 288$ in our framework. This is the same aspect ratio, \ie 10:3, used in the previous state-of-the-art UDA method ($640 \times 192$)~\cite{zheng2018t2net}.

At test time, we upsample the predictions ($960 \times 288$) to match the ground truth resolution ($1241 \times 376$) and apply the Garg cropping~\cite{garg2016unsupervised}. The reported results are based on the range of 1-80m for KITTI. Additionally, to show generalization performance on different scenes, we follow previous works for depth estimation~\cite{zheng2018t2net, zhou2017unsupervised, wang2018learning, monodepth2, johnston2020self} to test our model on the Make3D~\cite{make3d} outdoor scene dataset without re-training on the Make3D. For the evaluation on Make3D, we follow the same testing protocol and the evaluation criteria as in~\cite{monodepth2}.

We note that in contrast to \cite{zheng2018t2net}, we could not conduct an experiment on indoor scenes using SUNCG~\cite{song2016ssc} because SUNCG~\cite{song2016ssc} is no longer publicly available. Therefore, we follow other depth estimation works, such as \cite{monodepth17, monodepth2, zhao2019geometry}, to conduct extensive experiments on outdoor scenes using KITTI, the most used benchmark dataset for depth estimation, as well as vKITTI and Make3D.

%-------------------------------------------------------------------------

\begin{table*}[t]
\begin{center}
\scalebox{0.9}{
\begin{tabular}{l|ccccccc}
\hline
                            & \multicolumn{4}{c|}{Lower is better}                                     & \multicolumn{3}{c}{Higher is better}                       \\ \cline{2-8} 
\multirow{-2}{*}{Method}    & Abs Rel     & Sq Rel     & RMSE       & \multicolumn{1}{c|}{RMSE log}    & $\delta < 1.25$  & $\delta < 1.25^2$  & $\delta < 1.25^3$  \\ \hline
w/o SSRL                    & 0.171       & 1.423      & 5.781      & \multicolumn{1}{c|}{0.239}       & 0.776            & 0.916              & 0.968              \\
3rd last layer              & 0.173       & 1.376      & 5.570      & \multicolumn{1}{c|}{0.239}       & 0.768            & 0.914              & 0.968              \\
2nd last layer              & 0.170       & 1.318      & 5.558      & \multicolumn{1}{c|}{0.236}       & \bf 0.779        & \bf 0.924          & 0.972              \\
last layer                  & \bf 0.168   & \bf 1.228  & \bf 5.498  & \multicolumn{1}{c|}{\bf 0.235}   & 0.771            & 0.921              & \bf 0.973          \\ \hline
\end{tabular}
}
\end{center}
\caption{Ablation studies with different layers of $F_{dec}$ for the SSRL stage on KITTI.}
\label{ablation study feature}
\end{table*}

%-------------------------------------------------------------------------

\begin{table*}[t]
\begin{center}
\scalebox{0.9}{
\begin{tabular}{l|ccccccc}
\hline
                            & \multicolumn{4}{c|}{Lower is better}                                     & \multicolumn{3}{c}{Higher is better}                       \\ \cline{2-8} 
\multirow{-2}{*}{Method}    & Abs Rel     & Sq Rel     & RMSE       & \multicolumn{1}{c|}{RMSE log}    & $\delta < 1.25$  & $\delta < 1.25^2$  & $\delta < 1.25^3$  \\ \hline
global                      & 0.194       & 1.625      & 5.890      & \multicolumn{1}{c|}{0.248}       & 0.750            & 0.909              & 0.966 \\
pixel-wise                  & \bf 0.168   & \bf 1.228  & \bf 5.498  & \multicolumn{1}{c|}{\bf 0.235}   & \bf 0.771        & \bf 0.921          & \bf 0.973 \\ \hline
\end{tabular}
}
\end{center}
  \caption{Ablation studies with $L_{cossim}$ on KITTI.}
  \label{ablation study cossim}
\end{table*}

%-------------------------------------------------------------------------

\subsection{Implementation details}
We trained our framework using PyTorch on 8 NVIDIA A100 GPUs. During optimization, we set most of the the relative weights of the different loss functions based on previous works: $\lambda_{cycle}$ from ~\cite{CycleGAN2017}, and $\lambda_{task}$, $\lambda_{smooth}$, $\lambda_{identity}$ (or $\alpha_{r}$ in~\cite{zheng2018t2net}) from~\cite{zheng2018t2net}. We also conduct an ablation study on $\lambda_{crdoco}$ as we modify the task network. 
More specifically, we set the hyper-parameters as $\lambda_{cycle}=10$, $\lambda_{identity}=100$, $\lambda_{task}=100$, $\lambda_{smooth}=0.1$, $\lambda_{crdoco}=1$.

Our models are trained for 20 epochs (style transfer stage), 100 epochs (SSRL stage) and 30 epochs (depth estimation stage). We set the batch size of 16, 128, 16 for each stage and the learning rate of $lr\times Batch Size / 16$ (linear scaling~\cite{goyal2017accurate}), with a base $lr=0.0004$ for the task network $F$ and Siamese network $S$, and $lr=0.0002$ for the image translation network $G_{S\rightarrow T}$, $G_{T\rightarrow S}$, $D_{S}$, and $D_{T}$. Also, we applied a linearly decaying rate for the last half epochs in each stage. The training takes around 40 hours for the style transfer stage, 15 hours for the SSRL stage, and 40 hours for the depth estimation stage. For the depth estimation stage, we train our model twice and report the mean scores of all metrics. The implementation code and model weights will be made available.

\subsection{Comparisons}
\label{comparisons}
We use the standard metrics reported in~\cite{NIPS2014_7bccfde7} for evaluation. As comparison methods, we adopt the state-of-the-art UDA methods, T2Net~\cite{zheng2018t2net} and CrDoCo~\cite{CrDoCo}. We trained their models with their released code with the same resolution size as our model, \ie $960 \times 288$, for a fair comparison. 

In addition, as mentioned in Section~\ref{modifications}, we follow the recent standard to utilize ImageNet~\cite{imagenet_cvpr09} Initialization and design our encoder-decoder based task network with EfficientNetB5~\cite{tan2019efficientnet} pre-trained on ImageNet~\cite{imagenet_cvpr09} as an encoder. For a fair comparison, therefore, we implement the comparison methods, T2Net~\cite{zheng2018t2net} and CrDoCo~\cite{CrDoCo}, using our task network pre-trained on ImageNet~\cite{imagenet_cvpr09}.

Furthermore, unlike T2Net~\cite{zheng2018t2net} that prepares hyper-parameters specifically for outdoor scene depth estimation, CrDoCo~\cite{CrDoCo} did not conduct experiments on outdoor scenes and thus, their original hyper-parameters may not be suitable for outdoor scenes. This observation can be backed up with our experiment as shown in Table~\ref{vkitti2kitti} and a recent work~\cite{zhao2020domain}, which indicates that CrDoCo~\cite{CrDoCo} performs worse than T2Net~\cite{zheng2018t2net} in depth estimation tasks. In our work, we believe that using non-optimal hyper-parameters for comparison methods is less meaningful because we aim to not only achieve the state-of-the-art but also show the current exact performance gap between the UDA methods and state-of-the-art supervised counterparts. Therefore, we also report the result of CrDoCo~\cite{CrDoCo} using the same hyper-parameters as our model to show the full performance of the previous state-of-the-art method for a fairer comparison. We denote this model as CrDoCo$^{*}$ in Table~\ref{vkitti2kitti} and~\ref{make3d}.

Please note that we follow previous works~\cite{zhao2019geometry, zheng2018t2net} to include current SOTA depth estimation methods other than UDA as reference in Table~\ref{vkitti2kitti} and \ref{make3d}.

%-------------------------------------------------------------------------

\begin{table*}[t]
\begin{center}
\scalebox{0.9}{
\begin{tabular}{l|ccccccc}
\hline
                            & \multicolumn{4}{c|}{Lower is better}                                     & \multicolumn{3}{c}{Higher is better}                       \\ \cline{2-8} 
\multirow{-2}{*}{Method}    & Abs Rel     & Sq Rel     & RMSE       & \multicolumn{1}{c|}{RMSE log}    & $\delta < 1.25$  & $\delta < 1.25^2$  & $\delta < 1.25^3$  \\ \hline
$\alpha = 1$                & 0.185       & 1.596      & 5.964      & \multicolumn{1}{c|}{0.249}       & 0.757            & 0.910              & 0.968  \\
$\alpha = 2$                & 0.172       & 1.389      & 5.613      & \multicolumn{1}{c|}{0.238}       & \bf 0.771        & 0.915              & 0.969 \\
$\alpha = 3$                & \bf 0.168   & \bf 1.228  & \bf 5.498  & \multicolumn{1}{c|}{\bf 0.235}   & \bf 0.771        & \bf 0.921          & \bf 0.973 \\
$\alpha = 4$                & 0.170       & 1.306      & 5.528      & \multicolumn{1}{c|}{0.236}       & 0.768            & 0.918              & 0.971 \\ \hline
\end{tabular}
}
\end{center}
  \caption{Ablation studies with the different values of $\alpha$ in $S_{pre}$ on KITTI.}
  \label{ablation study hidden layer}
\end{table*}

%-------------------------------------------------------------------------

%-------------------------------------------------------------------------

\begin{table*}[t]
\begin{center}
\scalebox{0.9}{
\begin{tabular}{l|ccccccc}
\hline
                            & \multicolumn{4}{c|}{Lower is better}                                     & \multicolumn{3}{c}{Higher is better}                       \\ \cline{2-8} 
\multirow{-2}{*}{Method}    & Abs Rel     & Sq Rel     & RMSE       & \multicolumn{1}{c|}{RMSE log}    & $\delta < 1.25$  & $\delta < 1.25^2$  & $\delta < 1.25^3$  \\ \hline
    Task only w synthetic data & 0.176 & 1.752 & 6.209 & \multicolumn{1}{c|}{0.252} & 0.766 & 0.907 & 0.962 \\ 
    Task only w real data & 0.100 & 0.631 & 4.190 & \multicolumn{1}{c|}{0.165} & 0.886 & 0.968 & 0.988 \\ \cdashline{1-8}
    Ours & \bf 0.168 & \bf 1.228 & \bf 5.498 & \multicolumn{1}{c|}{\bf 0.235} & 0.771 & \bf 0.921 & \bf 0.973 \\ 
    (a) w SSRL and DE combined, w $L_{cossim}$ & 0.175 & 1.477 & 5.712 & \multicolumn{1}{c|}{0.240} & 0.769 & 0.913 & 0.967 \\
    (b) w SSRL and DE combined, w $S$ and $L_{cossim}$ & 0.171 & 1.302 & 5.520 & \multicolumn{1}{c|}{0.238} & \bf 0.773 & 0.915 & 0.969  \\
    (c) w local features for $S$ in SSRL & 0.189 & 1.621 & 5.841 & \multicolumn{1}{c|}{0.250} & 0.753 & 0.906 & 0.965 \\
    (d) w local features for $S_{pre}$ in SSRL & 0.174 & 1.463 & 5.799 & \multicolumn{1}{c|}{0.239} & 0.769 & 0.914 & 0.969 \\ \hline
\end{tabular}
}
\end{center}
  \caption{Ablation studies with variations of our framework on KITTI. Task: task network, S: Siamese network, SSRL: self-supervised representation learning stage, DE: depth estimation stage.}
  \label{ablation study variation}
\end{table*}

%-------------------------------------------------------------------------
\subsection{Results}
We present the results of depth estimation on KITTI~\cite{geiger2013vision} in Table~\ref{vkitti2kitti}. Our method outperforms previous state-of-the-art UDA methods~\cite{zheng2018t2net, CrDoCo} achieving constant improvement in all metrics (\eg 14.7\% on Sq Rel and 3.6\% on RMSE) and noticeably brings the performance closer to the fully supervised methods. See Fig.~\ref{fig:success} for a qualitative comparison. Our method is able to handle bright reflections and complex structures much better than other UDA methods, and thus, infer depths more reliably. Additional qualitative results are available in the supplementary material.

We provide the results on Make3D~\cite{make3d} in Table~\ref{make3d}. Again, our method performs better than all UDA methods on all metrics \eg by 16.9\% on Sq Rel. It is also noticeable that our method performs on par with the self-supervised counterparts that utilize extra information such as stereo or ego-motion~\cite{monodepth2, johnston2020self, NEURIPS2020_951124d4} and achieve the best result on log$_{10}$ among all of UDA and self-supervised methods. 

%-------------------------------------------------------------------------
\subsection{Ablation studies}
\label{sec:ablation}
We provide ablation studies on KITTI~\cite{geiger2013vision} in Table~\ref{ablation study feature}, ~\ref{ablation study cossim}, ~\ref{ablation study hidden layer}, and~\ref{ablation study variation}. Additional ablation studies are available in the supplementary material.

\paragraph{Layer of $F_{dec}$ for SSRL.} We study different layers of $F_{dec}$ for the SSRL stage, as summarized in Table~\ref{ablation study feature}. We first highlight that our framework even without our proposed SSRL stage (`w/o SSRL') achieves state-of-the-art performance against current SOTA methods in Table~\ref{vkitti2kitti}. In addition, we observe that using the representations for SSRL from the `last layer' or the `2nd to last layer' before the final output layer yields better performance than `w/o SSRL'. Note that the `2nd to last layer' model outperforms `w/o SSRL' on all metrics whereas the `last layer' model shows great improvement on almost all metrics with a slight drop in the $\delta < 1.25$ metric. For our experiments, we adopt `last layer' as mentioned in Section~\ref{SSRL} because we consider the improvement in Sq Rel (13.7\%) and RMSE (4.9\%) more significant than the slight decrease in $\delta < 1.25$ (0.6\%).

\paragraph{Cosine similarity loss $L_{cossim}$.} We study the effect of the cosine similarity loss $L_{cossim}$ in Table~\ref{ablation study cossim}. Specifically, we compare a global negative cosine similarity function, in which the correspondence is built between the whole feature maps, with the pixel-wise negative cosine similarity function as described in Section~\ref{SSRL}. The result indicates that `pixel-wise' performs better than `global' with significant margins for our high dimensional features, contributing to our state-of-the-art performance. This supports our assumption as described in Section~\ref{loss function in SSRL} that when the latent features are high dimensional, it is beneficial to account for the correspondence between the features at the level of pixels.

\paragraph{Scaling factor $\alpha$ for the predictor $S_{pre}$.} We explore the effect of different values of the scaling factor $\alpha$ that decides the number of the hidden layers' channel dimensions of $S_{pre}$ as shown in Table~\ref{ablation study hidden layer}. The result shows that relatively larger $\alpha$ (\ie the lower number of the channel dimensions) performs better than $\alpha = 1$ (\ie the same number of the channel dimensions as inputs). Thus, we observe the same tendency as indicated in~\cite{chen2020exploring} that `auto-encoder'-like structures can be effective for the predictor to digest information. Based on this result, we adopt $\alpha = 3$ for all experiments.

\paragraph{Variations of our framework.} We study different variations of our framework, as summarized in Table~\ref{ablation study variation}. 

Firstly, we train only the task network $F$ using either the synthetic or real dataset. It is worth noting that T2Net~\cite{zheng2018t2net} (Table~\ref{vkitti2kitti}) did not show better performance against the task network trained on only the synthetic dataset (`Task only w synthetic data'). We suspect that this is because T2Net~\cite{zheng2018t2net} is based on an end-to-end training with an image translation network and a task network. That is, at the beginning of the training, their translation network produces images of `bad quality' (\eg blurry images) that lead to inappropriate gradients for the encoder of the task network pre-traiend on ImangeNet. Meanwhile, our framework and CrDoCo*~\cite{CrDoCo} use a separate style transfer stage to pre-train the image translation network to produce images with 'good' quality in advance. Thus, these methods are able to make full use of the ImageNet initialization.

Secondly, we implement our framework with the SSRL and depth estimation stages combined to see if the separate SSRL stage is necessary. Specifically, we (a) apply only $L_{cossim}$ in the depth estimation stage or (b) utilize both the Siamese network $S$ and $L_{cossim}$ in the depth estimation stage. Although the latter model shows promising results, introducing the separate SSRL stage performs better. It is also worth noting that (a) performs worse than `w/o SSRL' in Table~\ref{ablation study feature}. We believe that this is because $L_{cossim}$ acts in a similar way as the cross-domain consistency loss $L_{crdoco}$~\cite{CrDoCo}, resulting in too strong prediction level consistency. This observation can be backed up by our ablation study on $L_{crdoco}$ in Table 8 in the supplementary material, which indicates that strong prediction level consistency deteriorates the performance. 

Thirdly, we explore ways to input feature maps into $S$ instead of using global feature maps as inputs to $S$ as described in Section~\ref{SSRL}. One way is to divide the feature maps into several local feature blocks, which are forwarded to $S$. Specifically, (c) we separate $M$ of the size $12 \times 960 \times 288$ into 30 blocks of the size $12 \times 96 \times 96$, and input the blocks into $S$. In another way, (d) we perform the same operation to the output of $S_{pro}$ and apply them to only $S_{pre}$. As shown in Table~\ref{ablation study variation}, however, using global features yields the best performance; therefore, we adopt it for all experiments.

%-------------------------------------------------------------------------

\section{Conclusion}
\label{conclusion}
We introduced a novel framework for unsupervised domain adaptation for monocular depth estimation. We propose using a bidirectional image translation and a Siamese network to learn representations that are invariant across real and synthetic domains in a self-supervised manner. Our extensive results demonstrate that our method brings decisive improvements both quantitatively and qualitatively on two popular datasets, KITTI and Make3D, \eg by 14.7\% for KITTI and 16.9\% for Make3D on square relative error (Sq Rel). Limitation: our method requires an image-to-image translation network to provide two style augmented views of the same scene.  In future work, we would like to explore modifications that do not require paired style augmented images. We would also like to investigate how our framework generalizes to other dense prediction tasks such as semantic segmentation.

{\small
\bibliographystyle{ieee_fullname}
\bibliography{main}

\begin{thebibliography}{10}\itemsep=-1pt

\bibitem{Alhashim2018}
Ibraheem Alhashim and Peter Wonka.
\newblock High quality monocular depth estimation via transfer learning.
\newblock {\em arXiv e-prints}, 2018.

\bibitem{make3d}
Andrew Y.~Ng. Ashutosh~Saxena, Min~Sun.
\newblock Make3d: Learning 3d scene structure from a single still image.
\newblock {\em IEEE Transactions of Pattern Analysis and Machine Intelligence
  (PAMI)}, 30(5):824--840, 2009.

\bibitem{bhat2020adabins}
Shariq~Farooq Bhat, Ibraheem Alhashim, and Peter Wonka.
\newblock Adabins: Depth estimation using adaptive bins, 2020.

\bibitem{Bi_2019_ICCV}
Sai Bi, Kalyan Sunkavalli, Federico Perazzi, Eli Shechtman, Vladimir~G. Kim,
  and Ravi Ramamoorthi.
\newblock Deep cg2real: Synthetic-to-real translation via image
  disentanglement.
\newblock In {\em Proceedings of the IEEE/CVF International Conference on
  Computer Vision (ICCV)}, October 2019.

\bibitem{chen2019domain}
Minghao Chen, Hongyang Xue, and Deng Cai.
\newblock Domain adaptation for semantic segmentation with maximum squares
  loss.
\newblock In {\em Proceedings of the IEEE/CVF International Conference on
  Computer Vision}, pages 2090--2099, 2019.

\bibitem{chen2020simple}
Ting Chen, Simon Kornblith, Mohammad Norouzi, and Geoffrey Hinton.
\newblock A simple framework for contrastive learning of visual
  representations.
\newblock In {\em International conference on machine learning}, pages
  1597--1607. PMLR, 2020.

\bibitem{NEURIPS2020_fcbc95cc}
Ting Chen, Simon Kornblith, Kevin Swersky, Mohammad Norouzi, and Geoffrey~E
  Hinton.
\newblock Big self-supervised models are strong semi-supervised learners.
\newblock In H. Larochelle, M. Ranzato, R. Hadsell, M.~F. Balcan, and H. Lin,
  editors, {\em Advances in Neural Information Processing Systems}, volume~33,
  pages 22243--22255. Curran Associates, Inc., 2020.

\bibitem{chen2021contrastive}
Wuyang Chen, Zhiding Yu, SD Mello, Sifei Liu, Jose~M Alvarez, Zhangyang Wang,
  and Anima Anandkumar.
\newblock Contrastive syn-to-real generalization.
\newblock 2021.

\bibitem{chen2020improved}
Xinlei Chen, Haoqi Fan, Ross Girshick, and Kaiming He.
\newblock Improved baselines with momentum contrastive learning.
\newblock {\em arXiv preprint arXiv:2003.04297}, 2020.

\bibitem{chen2020exploring}
Xinlei Chen and Kaiming He.
\newblock Exploring simple siamese representation learning.
\newblock In {\em Proceedings of the IEEE/CVF Conference on Computer Vision and
  Pattern Recognition}, pages 15750--15758, 2021.

\bibitem{chen2021empirical}
Xinlei Chen, Saining Xie, and Kaiming He.
\newblock An empirical study of training self-supervised visual transformers.
\newblock {\em arXiv preprint arXiv:2104.02057}, 2021.

\bibitem{CrDoCo}
Yun-Chun Chen, Yen-Yu Lin, Ming-Hsuan Yang, and Jia-Bin Huang.
\newblock Crdoco: Pixel-level domain transfer with cross-domain consistency.
\newblock In {\em IEEE Conference on Computer Vision and Pattern Recognition
  (CVPR)}, 2019.

\bibitem{imagenet_cvpr09}
J. Deng, W. Dong, R. Socher, L.-J. Li, K. Li, and L. Fei-Fei.
\newblock {ImageNet: A Large-Scale Hierarchical Image Database}.
\newblock In {\em CVPR09}, 2009.

\bibitem{NIPS2014_7bccfde7}
David Eigen, Christian Puhrsch, and Rob Fergus.
\newblock Depth map prediction from a single image using a multi-scale deep
  network.
\newblock In Z. Ghahramani, M. Welling, C. Cortes, N. Lawrence, and K.~Q.
  Weinberger, editors, {\em Advances in Neural Information Processing Systems},
  volume~27. Curran Associates, Inc., 2014.

\bibitem{fu2018deep}
Huan Fu, Mingming Gong, Chaohui Wang, Kayhan Batmanghelich, and Dacheng Tao.
\newblock Deep ordinal regression network for monocular depth estimation.
\newblock In {\em Proceedings of the IEEE Conference on Computer Vision and
  Pattern Recognition}, pages 2002--2011, 2018.

\bibitem{gaidon2016virtual}
Adrien Gaidon, Qiao Wang, Yohann Cabon, and Eleonora Vig.
\newblock Virtual worlds as proxy for multi-object tracking analysis.
\newblock In {\em Proceedings of the IEEE conference on computer vision and
  pattern recognition}, pages 4340--4349, 2016.

\bibitem{gan2018monocular}
Yukang Gan, Xiangyu Xu, Wenxiu Sun, and Liang Lin.
\newblock Monocular depth estimation with affinity, vertical pooling, and label
  enhancement.
\newblock In {\em Proceedings of the European Conference on Computer Vision
  (ECCV)}, pages 224--239, 2018.

\bibitem{garg2016unsupervised}
Ravi Garg, Vijay~Kumar Bg, Gustavo Carneiro, and Ian Reid.
\newblock Unsupervised cnn for single view depth estimation: Geometry to the
  rescue.
\newblock In {\em European conference on computer vision}, pages 740--756.
  Springer, 2016.

\bibitem{geiger2013vision}
Andreas Geiger, Philip Lenz, Christoph Stiller, and Raquel Urtasun.
\newblock Vision meets robotics: The kitti dataset.
\newblock {\em The International Journal of Robotics Research},
  32(11):1231--1237, 2013.

\bibitem{monodepth17}
Cl{\'{e}}ment Godard, Oisin {Mac Aodha}, and Gabriel~J. Brostow.
\newblock Unsupervised monocular depth estimation with left-right consistency.
\newblock In {\em CVPR}, 2017.

\bibitem{godard2017unsupervised}
Cl{\'e}ment Godard, Oisin Mac~Aodha, and Gabriel~J Brostow.
\newblock Unsupervised monocular depth estimation with left-right consistency.
\newblock In {\em Proceedings of the IEEE Conference on Computer Vision and
  Pattern Recognition}, pages 270--279, 2017.

\bibitem{monodepth2}
Cl{\'e}ment Godard, Oisin Mac~Aodha, Michael Firman, and Gabriel~J Brostow.
\newblock Digging into self-supervised monocular depth estimation.
\newblock In {\em Proceedings of the IEEE/CVF International Conference on
  Computer Vision}, pages 3828--3838, 2019.

\bibitem{NEURIPS2020_951124d4}
Juan~Luis GonzalezBello and Munchurl Kim.
\newblock Forget about the lidar: Self-supervised depth estimators with med
  probability volumes.
\newblock In H. Larochelle, M. Ranzato, R. Hadsell, M.~F. Balcan, and H. Lin,
  editors, {\em Advances in Neural Information Processing Systems}, volume~33,
  pages 12626--12637. Curran Associates, Inc., 2020.

\bibitem{goodfellow2014generative}
Ian~J Goodfellow, Jean Pouget-Abadie, Mehdi Mirza, Bing Xu, David Warde-Farley,
  Sherjil Ozair, Aaron Courville, and Yoshua Bengio.
\newblock Generative adversarial networks.
\newblock {\em arXiv preprint arXiv:1406.2661}, 2014.

\bibitem{goyal2017accurate}
Priya Goyal, Piotr Doll{\'a}r, Ross Girshick, Pieter Noordhuis, Lukasz
  Wesolowski, Aapo Kyrola, Andrew Tulloch, Yangqing Jia, and Kaiming He.
\newblock Accurate, large minibatch sgd: Training imagenet in 1 hour.
\newblock {\em arXiv preprint arXiv:1706.02677}, 2017.

\bibitem{NEURIPS2020_f3ada80d}
Jean-Bastien Grill, Florian Strub, Florent Altch\'{e}, Corentin Tallec, Pierre
  Richemond, Elena Buchatskaya, Carl Doersch, Bernardo Avila~Pires, Zhaohan
  Guo, Mohammad Gheshlaghi~Azar, Bilal Piot, Koray Kavukcuoglu, Remi Munos, and
  Michal Valko.
\newblock Bootstrap your own latent - a new approach to self-supervised
  learning.
\newblock In H. Larochelle, M. Ranzato, R. Hadsell, M.~F. Balcan, and H. Lin,
  editors, {\em Advances in Neural Information Processing Systems}, volume~33,
  pages 21271--21284. Curran Associates, Inc., 2020.

\bibitem{hadsell2006dimensionality}
Raia Hadsell, Sumit Chopra, and Yann LeCun.
\newblock Dimensionality reduction by learning an invariant mapping.
\newblock In {\em 2006 IEEE Computer Society Conference on Computer Vision and
  Pattern Recognition (CVPR'06)}, volume~2, pages 1735--1742. IEEE, 2006.

\bibitem{he2020momentum}
Kaiming He, Haoqi Fan, Yuxin Wu, Saining Xie, and Ross Girshick.
\newblock Momentum contrast for unsupervised visual representation learning.
\newblock In {\em Proceedings of the IEEE/CVF Conference on Computer Vision and
  Pattern Recognition}, pages 9729--9738, 2020.

\bibitem{heise2013pm}
Philipp Heise, Sebastian Klose, Brian Jensen, and Alois Knoll.
\newblock Pm-huber: Patchmatch with huber regularization for stereo matching.
\newblock In {\em Proceedings of the IEEE International Conference on Computer
  Vision}, pages 2360--2367, 2013.

\bibitem{ioffe2015batch}
Sergey Ioffe and Christian Szegedy.
\newblock Batch normalization: Accelerating deep network training by reducing
  internal covariate shift.
\newblock In {\em International conference on machine learning}, pages
  448--456. PMLR, 2015.

\bibitem{johnston2020self}
Adrian Johnston and Gustavo Carneiro.
\newblock Self-supervised monocular trained depth estimation using
  self-attention and discrete disparity volume.
\newblock In {\em Proceedings of the IEEE/CVF Conference on Computer Vision and
  Pattern Recognition}, pages 4756--4765, 2020.

\bibitem{kundu2018adadepth}
Jogendra~Nath Kundu, Phani~Krishna Uppala, Anuj Pahuja, and R~Venkatesh Babu.
\newblock Adadepth: Unsupervised content congruent adaptation for depth
  estimation.
\newblock In {\em Proceedings of the IEEE conference on computer vision and
  pattern recognition}, pages 2656--2665, 2018.

\bibitem{kuznietsov2017semi}
Yevhen Kuznietsov, Jorg Stuckler, and Bastian Leibe.
\newblock Semi-supervised deep learning for monocular depth map prediction.
\newblock In {\em Proceedings of the IEEE conference on computer vision and
  pattern recognition}, pages 6647--6655, 2017.

\bibitem{lee2019big}
Jin~Han Lee, Myung-Kyu Han, Dong~Wook Ko, and Il~Hong Suh.
\newblock From big to small: Multi-scale local planar guidance for monocular
  depth estimation.
\newblock {\em arXiv preprint arXiv:1907.10326}, 2019.

\bibitem{li2020content}
Guangrui Li, Guoliang Kang, Wu Liu, Yunchao Wei, and Yi Yang.
\newblock Content-consistent matching for domain adaptive semantic
  segmentation.
\newblock In {\em European Conference on Computer Vision}, pages 440--456.
  Springer, 2020.

\bibitem{lian2019constructing}
Qing Lian, Fengmao Lv, Lixin Duan, and Boqing Gong.
\newblock Constructing self-motivated pyramid curriculums for cross-domain
  semantic segmentation: A non-adversarial approach.
\newblock In {\em Proceedings of the IEEE/CVF International Conference on
  Computer Vision}, pages 6758--6767, 2019.

\bibitem{liu2021domain}
Weizhe Liu, David Ferstl, Samuel Schulter, Lukas Zebedin, Pascal Fua, and
  Christian Leistner.
\newblock Domain adaptation for semantic segmentation via patch-wise
  contrastive learning, 2021.

\bibitem{lopez2020desc}
Adrian Lopez-Rodriguez and Krystian Mikolajczyk.
\newblock Desc: Domain adaptation for depth estimation via semantic
  consistency.
\newblock In {\em British Machine Vision Conference (BMVC)}, 2020.

\bibitem{lv2020cross}
Fengmao Lv, Tao Liang, Xiang Chen, and Guosheng Lin.
\newblock Cross-domain semantic segmentation via domain-invariant interactive
  relation transfer.
\newblock In {\em Proceedings of the IEEE/CVF Conference on Computer Vision and
  Pattern Recognition}, pages 4334--4343, 2020.

\bibitem{mao2016multi}
Xudong Mao, Qing Li, Haoran Xie, Raymond~YK Lau, and Zhen Wang.
\newblock Multi-class generative adversarial networks with the l2 loss
  function.
\newblock {\em arXiv preprint arXiv:1611.04076}, 5:00102, 2016.

\bibitem{DBLP:journals/corr/abs-1807-10915}
Andrea Pilzer, Dan Xu, Mihai~Marian Puscas, Elisa Ricci, and Nicu Sebe.
\newblock Unsupervised adversarial depth estimation using cycled generative
  networks.
\newblock {\em CoRR}, abs/1807.10915, 2018.

\bibitem{pnvr2020sharingan}
Koutilya PNVR, Hao Zhou, and David Jacobs.
\newblock Sharingan: Combining synthetic and real data for unsupervised
  geometry estimation.
\newblock In {\em Proceedings of the IEEE/CVF Conference on Computer Vision and
  Pattern Recognition}, pages 13974--13983, 2020.

\bibitem{ronneberger2015u}
Olaf Ronneberger, Philipp Fischer, and Thomas Brox.
\newblock U-net: Convolutional networks for biomedical image segmentation.
\newblock In {\em International Conference on Medical image computing and
  computer-assisted intervention}, pages 234--241. Springer, 2015.

\bibitem{sankaranarayanan2018learning}
Swami Sankaranarayanan, Yogesh Balaji, Arpit Jain, Ser~Nam Lim, and Rama
  Chellappa.
\newblock Learning from synthetic data: Addressing domain shift for semantic
  segmentation.
\newblock In {\em Proceedings of the IEEE Conference on Computer Vision and
  Pattern Recognition}, pages 3752--3761, 2018.

\bibitem{song2016ssc}
Shuran Song, Fisher Yu, Andy Zeng, Angel~X Chang, Manolis Savva, and Thomas
  Funkhouser.
\newblock Semantic scene completion from a single depth image.
\newblock {\em Proceedings of 30th IEEE Conference on Computer Vision and
  Pattern Recognition}, 2017.

\bibitem{subhani2020learning}
M~Naseer Subhani and Mohsen Ali.
\newblock Learning from scale-invariant examples for domain adaptation in
  semantic segmentation.
\newblock {\em arXiv preprint arXiv:2007.14449}, 2020.

\bibitem{taigman2016unsupervised}
Yaniv Taigman, Adam Polyak, and Lior Wolf.
\newblock Unsupervised cross-domain image generation.
\newblock {\em arXiv preprint arXiv:1611.02200}, 2016.

\bibitem{tan2019efficientnet}
Mingxing Tan and Quoc Le.
\newblock Efficientnet: Rethinking model scaling for convolutional neural
  networks.
\newblock In {\em International Conference on Machine Learning}, pages
  6105--6114. PMLR, 2019.

\bibitem{DBLP:journals/corr/abs-1909-03943}
Alessio Tonioni, Matteo Poggi, Stefano Mattoccia, and Luigi di Stefano.
\newblock Unsupervised domain adaptation for depth prediction from images.
\newblock {\em CoRR}, abs/1909.03943, 2019.

\bibitem{wang2018learning}
Chaoyang Wang, Jos{\'e}~Miguel Buenaposada, Rui Zhu, and Simon Lucey.
\newblock Learning depth from monocular videos using direct methods.
\newblock In {\em Proceedings of the IEEE Conference on Computer Vision and
  Pattern Recognition}, pages 2022--2030, 2018.

\bibitem{wu2018dcan}
Zuxuan Wu, Xintong Han, Yen-Liang Lin, Mustafa~Gokhan Uzunbas, Tom Goldstein,
  Ser~Nam Lim, and Larry~S Davis.
\newblock Dcan: Dual channel-wise alignment networks for unsupervised scene
  adaptation.
\newblock In {\em Proceedings of the European Conference on Computer Vision
  (ECCV)}, pages 518--534, 2018.

\bibitem{xie2020propagate}
Zhenda Xie, Yutong Lin, Zheng Zhang, Yue Cao, Stephen Lin, and Han Hu.
\newblock Propagate yourself: Exploring pixel-level consistency for
  unsupervised visual representation learning.
\newblock 2021.

\bibitem{yang2020fda}
Yanchao Yang and Stefano Soatto.
\newblock Fda: Fourier domain adaptation for semantic segmentation.
\newblock In {\em Proceedings of the IEEE/CVF Conference on Computer Vision and
  Pattern Recognition}, pages 4085--4095, 2020.

\bibitem{DualGAN}
Z. {Yi}, H. {Zhang}, P. {Tan}, and M. {Gong}.
\newblock Dualgan: Unsupervised dual learning for image-to-image translation.
\newblock In {\em 2017 IEEE International Conference on Computer Vision
  (ICCV)}, pages 2868--2876, 2017.

\bibitem{yin2019enforcing}
Wei Yin, Yifan Liu, Chunhua Shen, and Youliang Yan.
\newblock Enforcing geometric constraints of virtual normal for depth
  prediction.
\newblock In {\em Proceedings of the IEEE/CVF International Conference on
  Computer Vision}, pages 5684--5693, 2019.

\bibitem{Zhang_2020_CVPR}
Zhenyu Zhang, Stephane Lathuiliere, Elisa Ricci, Nicu Sebe, Yan Yan, and Jian
  Yang.
\newblock Online depth learning against forgetting in monocular videos.
\newblock In {\em Proceedings of the IEEE/CVF Conference on Computer Vision and
  Pattern Recognition (CVPR)}, June 2020.

\bibitem{zhao2019geometry}
Shanshan Zhao, Huan Fu, Mingming Gong, and Dacheng Tao.
\newblock Geometry-aware symmetric domain adaptation for monocular depth
  estimation.
\newblock In {\em Proceedings of the IEEE/CVF Conference on Computer Vision and
  Pattern Recognition}, pages 9788--9798, 2019.

\bibitem{zhao2020domain}
Yunhan Zhao, Shu Kong, Daeyun Shin, and Charless Fowlkes.
\newblock Domain decluttering: Simplifying images to mitigate synthetic-real
  domain shift and improve depth estimation.
\newblock In {\em Proceedings of the IEEE/CVF Conference on Computer Vision and
  Pattern Recognition}, pages 3330--3340, 2020.

\bibitem{zheng2018t2net}
Chuanxia Zheng, Tat-Jen Cham, and Jianfei Cai.
\newblock T2net: Synthetic-to-realistic translation for solving single-image
  depth estimation tasks.
\newblock In {\em Proceedings of the European Conference on Computer Vision
  (ECCV)}, pages 767--783, 2018.

\bibitem{zhou2017unsupervised}
Tinghui Zhou, Matthew Brown, Noah Snavely, and David~G Lowe.
\newblock Unsupervised learning of depth and ego-motion from video.
\newblock In {\em Proceedings of the IEEE conference on computer vision and
  pattern recognition}, pages 1851--1858, 2017.

\bibitem{CycleGAN2017}
Jun-Yan Zhu, Taesung Park, Phillip Isola, and Alexei~A Efros.
\newblock Unpaired image-to-image translation using cycle-consistent
  adversarial networks.
\newblock In {\em Computer Vision (ICCV), 2017 IEEE International Conference
  on}, 2017.

\bibitem{zhu2018penalizing}
Xinge Zhu, Hui Zhou, Ceyuan Yang, Jianping Shi, and Dahua Lin.
\newblock Penalizing top performers: Conservative loss for semantic
  segmentation adaptation.
\newblock In {\em Proceedings of the European Conference on Computer Vision
  (ECCV)}, pages 568--583, 2018.

\bibitem{zou2018domain}
Yang Zou, Zhiding Yu, BVK Kumar, and Jinsong Wang.
\newblock Domain adaptation for semantic segmentation via class-balanced
  self-training.
\newblock {\em arXiv preprint arXiv:1810.07911}, 2018.

\end{thebibliography}
}

\clearpage

\appendix

%%%%%%%%% BODY TEXT

\section{Training stages and loss functions}
In this section, we describe the style transfer stage, the depth estimation stage, and relevant loss functions in more details. Please also see Section 3 in our main paper.

\subsection{Style transfer stage}
In this stage, we aim to train the image-to-image translation networks $G_{S\rightarrow T}$ and $G_{T\rightarrow S}$. $G_{S\rightarrow T}$ learns a mapping between the synthetic (\ie source) and realistic (\ie target) domain. $G_{T\rightarrow S}$ learns a mapping in the opposite direction. Previous work~\cite{zheng2018t2net, zhao2019geometry} 
has shown that 
% standalone training of the image translation networks may not be optimal for the synthetic-to-realistic image translation task due to the more artifacts generated by such a network, and 
the depth supervision can help to improve the quality of style transfer compared to the standalone training of the translation networks. Although the effect of the depth supervision seems minor based on our experiment as shown in Table~\ref{depth supervision}, we adopt it for all experiments and jointly train the image-to-image translation networks $G_{S\rightarrow T}$ and $G_{T\rightarrow S}$, and the image-to-depth task network $F$ with synthetic depth labels. Here, based on previous work~\cite{zheng2018t2net, CrDoCo}, we adopt six loss functions: adversarial loss $L_{adv}$, cycle consistency loss $L_{cycle}$, identity mapping loss $L_{identity}$ (or reconstruction loss in~\cite{zheng2018t2net}), task loss $L_{task}$, smooth loss $L_{smooth}$, and cross domain consistency loss $L_{crdoco}$. These losses are described in the following.

%-------------------------------------------------------------------------

\paragraph{Adversarial loss} We utilize adversarial training for the image translation. Following CycleGAN~\cite{CycleGAN2017}, we employ two generators $G_{S\rightarrow T}$ and $G_{T\rightarrow S}$, and two discriminators $D_{s}$ and $D_{t}$ for the source and target domains, respectively. $G_{S\rightarrow T}$ tries to learn the mapping from the source to the target domain, \ie $G_{S\rightarrow T}: I_{S}\rightarrow I_{S\rightarrow T}$, such that the data distribution of the translated images from the source domain $I_{S\rightarrow T}$ is indistinguishable from that of the target domain $I_{T}$. Then, $D_{T}$ aims to distinguish between the images in the target domain $I_{T}$ and the translated images from the source domain $I_{S\rightarrow T}$. Thus, using the technique of a least-square loss~\cite{mao2016multi} for stable training, we define adversarial loss~\cite{goodfellow2014generative} in the target domain as 

\begin{equation}
\label{adversarial loss}
\begin{split}
 L_{adv}&(G_{S\rightarrow T}, D_T, X_S, X_T) \\
 &= \mathbb{E}_{I_T \sim X_T}[(D_T(I_T) - 1)^2] \\
 &+ \mathbb{E}_{I_S \sim X_S}[(D_T(G_{S\rightarrow T}(I_S)))^2].
\end{split} 
\end{equation}

Similarly, the adversarial loss for the mapping function $G_{T\rightarrow S}: I_T \rightarrow I_{T\rightarrow S}$ is introduced as $L_{adv}(G_{T\rightarrow S}, D_S, X_T, X_S)$. 
Please note that as a result of the bidirectional adversarial training, we obtain two labeled types of images $I_S$ and $I_{S\rightarrow T}$, and two unlabeled types of images $I_T$ and $I_{T\rightarrow S}$ as shown in Fig 2 in our main paper.

%-------------------------------------------------------------------------

\paragraph{Cycle consistency loss $L_{cycle}$.} Generally, training $G_{S\rightarrow T}$ and $G_{T\rightarrow S}$ with only the adversarial loss are highly under-constrained. To further regularize the translation network, we use a cycle consistency loss \cite{CycleGAN2017, DualGAN}. This loss function is based on the idea that when images are translated from one domain to another, followed by an inverse translation, the reconstructed images should be the same as the original, \ie $G_{T\rightarrow S}(G_{S\rightarrow T}(I_S)) \approx I_S$ and $G_{S\rightarrow T}(G_{T\rightarrow S}(I_T)) \approx I_T$. Therefore, we define the cycle consistency loss as 

\begin{equation}
\label{cycle consistency loss}
\begin{split}
 L_{cycle}&(G_{S\rightarrow T}, G_{T\rightarrow S}, X_S, X_T) \\
 &= \mathbb{E}_{I_S \sim X_S}[\|G_{T\rightarrow S}(G_{S\rightarrow T}(I_S))) - I_S\|_1] \\
 &+ \mathbb{E}_{I_T \sim X_T}[\|G_{S\rightarrow T}(G_{T\rightarrow S}(I_T)) - I_T\|_1].
\end{split} 
\end{equation}

%-------------------------------------------------------------------------

\begin{table*}[t]
\begin{center}
\scalebox{1}{
\begin{tabular}{l|ccccccc}
\hline
                                                                                                                                                                                                                       & \multicolumn{4}{c|}{Lower is better}                                                                   & \multicolumn{3}{c}{Higher is better}                               \\ \cline{2-8} 
\multicolumn{1}{l|}{\multirow{-2}{*}{Method}}                                                              & Abs Rel              & Sq Rel               & RMSE                 & \multicolumn{1}{c|}{RMSE log}     & $\delta < 1.25$      & $\delta < 1.25^2$    & $\delta < 1.25^3$    \\ \hline
\multicolumn{1}{l|}{No depth supervision}                                                                                                                                                                 & \bf 0.174                & 1.490                & 5.751                & \multicolumn{1}{c|}{0.244}        & 0.766                & 0.910                & 0.964                \\
\multicolumn{1}{l|}{depth supervision}                                                                                                                                                                    & \bf 0.174                & \bf 1.439                & \bf 5.701                & \multicolumn{1}{c|}{\bf 0.241}        & \bf 0.770                & \bf 0.914                & \bf 0.967                \\ \hline
\end{tabular}
}
\end{center}
\caption{Ablation studies on the effect of the depth supervision in the style transfer stage using CrDoCo*~\cite{CrDoCo} on KITTI~\cite{geiger2013vision}.}
\label{depth supervision}
\end{table*}

\begin{table*}[t]
\begin{center}
\scalebox{1}{
\begin{tabular}{l|ccccccc}
\hline
                                                                                                                                                                                                                       & \multicolumn{4}{c|}{Lower is better}                                                                   & \multicolumn{3}{c}{Higher is better}                               \\ \cline{2-8} 
\multirow{-2}{*}{Method}                                                              & Abs Rel              & Sq Rel               & RMSE                 & \multicolumn{1}{c|}{RMSE log}     & $\delta < 1.25$      & $\delta < 1.25^2$    & $\delta < 1.25^3$    \\ \hline
$\lambda_{crdoco} = 10.0$ & 0.178 & 1.269 & 5.845 & \multicolumn{1}{c|}{0.245} & 0.742 & 0.910 & 0.970 \\
$\lambda_{crdoco} = 1.0$ & \bf 0.168 & \bf 1.228 & \bf 5.498 & \multicolumn{1}{c|}{\bf 0.235} & \bf 0.771 & \bf 0.921 & \bf 0.973 \\
w/o $\lambda_{crdoco}$ & 0.176 & 1.443 & 5.676 & \multicolumn{1}{c|}{0.240} & 0.768 & 0.917 & 0.971 \\ \hline
\end{tabular}
}
\end{center}
  \caption{Ablation studies with the different number of $\lambda_{crdoco}$ on KITTI~\cite{geiger2013vision}.}
  \label{ablation study crdoco}
\end{table*}

\paragraph{Identity mapping loss $L_{identity}$.} In addition to $L_{cycle}$, we also use an identity mapping loss \cite{CycleGAN2017, taigman2016unsupervised} to regularize the training of $G_{S\rightarrow T}$ and $G_{T\rightarrow S}$. Note that in~\cite{zheng2018t2net}, this loss function is named as `reconstruction loss'. The identity mapping loss encourages $G_{S\rightarrow T}$ and $G_{T\rightarrow S}$ to preserve image styles when the input images already belong to the translation-target domain,
% belong to the same domain that each translation network aims at,
\ie $G_{S\rightarrow T}(I_T)\approx I_T$ and $G_{T\rightarrow S}(I_S)\approx I_S$. Specifically, we define the identity mapping loss as 

\begin{equation}
\label{identity mapping loss}
\begin{split}
 L_{identity}&(G_{S\rightarrow T}, G_{T\rightarrow S}, X_S, X_T) \\
 &= \mathbb{E}_{I_S \sim X_S}[\|G_{T\rightarrow S}(I_S)) - I_S\|_1] \\
 &+ \mathbb{E}_{I_T \sim X_T}[\|G_{S\rightarrow T}(I_T) - I_T\|_1].
\end{split} 
\end{equation}
%-------------------------------------------------------------------------

\paragraph{Task loss $L_{task}$.} To train the image-to-depth task network $F$, we provide supervision to $F$ using synthetic ground truth depth maps $I_{S, lab}$. Specifically, we pass the two labeled types of images $I_S$ and $I_{S\rightarrow T}$ to $F$ to obtain corresponding depth maps $P_{S} = F(I_{S})$ and $P_{S\rightarrow T} = F(I_{S\rightarrow T})$ as shown in Fig 1 in our main paper. Since these depth maps should have the same label $I_{S,lab}$, we define the task loss as 

\begin{equation}
\label{task loss}
\begin{split}
 L_{task}&(G_{S\rightarrow T}, F, X_S) \\
 &= \mathbb{E}_{I_S \sim X_S}[\|F(I_S) - I_{S,lab}\|_1] \\
 &+ \mathbb{E}_{I_S \sim X_S}[\|F(G_{S\rightarrow T}(I_S)) - I_{S, lab}\|_1].
\end{split} 
\end{equation}

%-------------------------------------------------------------------------

\paragraph{Smooth loss $L_{smooth}$.} Following previous works~\cite{zheng2018t2net, garg2016unsupervised, godard2017unsupervised, heise2013pm, kuznietsov2017semi, zhao2019geometry, CrDoCo} we utilize a a smooth loss to guide a more reasonable depth estimation using the unlabeled images $I_T$ and $I_{S\rightarrow T}$. Specifically, we use a robust penalty with an edge-aware term for $I_T$ as 

\begin{equation}
\label{smooth loss}
\begin{split}
 L_{smooth}(F, X_T) &= \mathbb{E}_{I_T \sim X_T}[|\partial _x F(I_T)|e^{-|\partial _x I_T|}]\\ 
 &+ \mathbb{E}_{I_T \sim X_T}[|\partial _y F(I_T)|e^{-|\partial _y I_T|}].
\end{split} 
\end{equation}

Similarly, the smooth loss for $I_{S\rightarrow T}$ is introduced, \ie $L_{smooth}(F, X_{S\rightarrow T})$.

%-------------------------------------------------------------------------

\begin{figure*}[t]
  \centering
  \begin{tabular}{cc}
  \begin{minipage}[c]{0.00001\hsize}
    \caption*{\rotatebox{90}{{\scriptsize \,\,\,\,\,\, Ours \,\,\,\,\,\, CrDoCo*~\cite{CrDoCo} \, T$2$Net~\cite{zheng2018t2net} \,\,\,\,\,\,\,\, input \,\,\,\,}}}
  \end{minipage} &
  \begin{minipage}[c]{0.95\hsize}
  \includegraphics[width=1.0\linewidth]{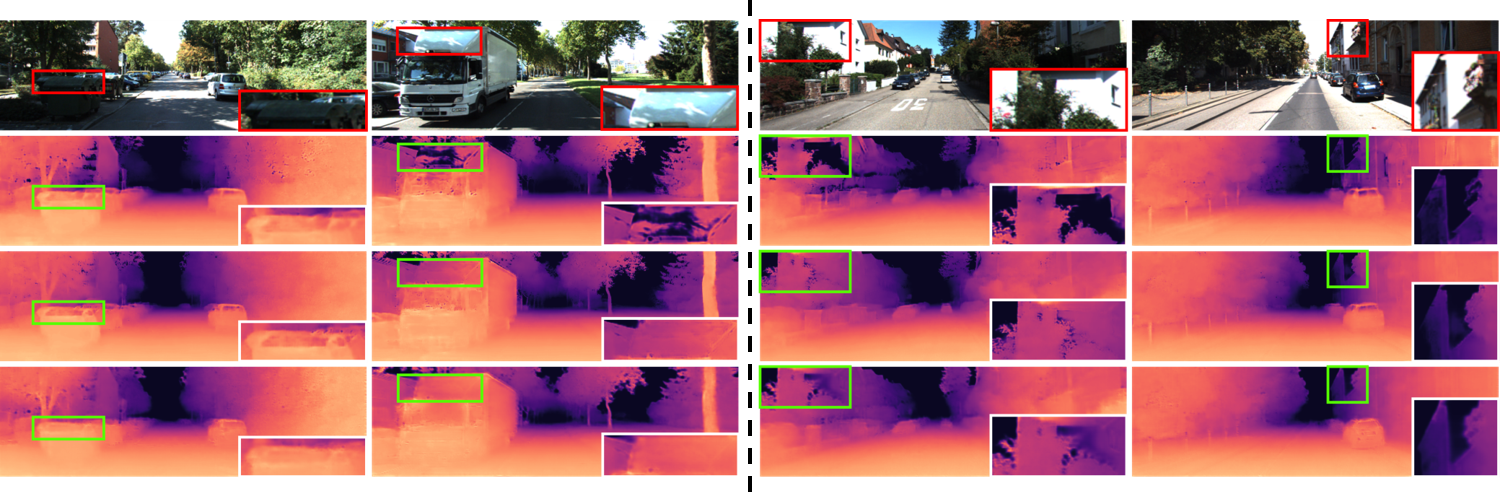}\\
  \end{minipage}
  \end{tabular}
   \vspace{-15pt}
    \caption*{Successful case \,\,\,\,\,\,\,\,\,\,\,\,\,\,\,\,\,\,\,\,\,\,\,\,\,\,\,\,\,\,\,\,\,\,\,\,\,\,\,\,\,\,\,\,\,\,\,\,\,\,\,\,\,\,\,\,\,\,\,\,\,\,\,\,\,\,\,\,\,\,\,\,\,\,\,\,\,\,\,\,\,\,\,\,\,\,\,\,\,\,\,\,\,\,\,\,\,\,\,\,\,\,\,\,\,\,\,\,\, Failure case}
  \vspace{5pt}
  \caption{Qualitative results for successful and failure cases on KITTI~\cite{geiger2013vision}.}
  \label{fig:additional qualitative results}
\end{figure*}

\paragraph{Cross domain consistency loss $L_{crdoco}$.} Following the previous work~\cite{CrDoCo}, we also introduce a cross domain consistency loss to enforce the consistency between the depth predictions of the two unlabeled types of images $P_{T} = F(I_{T})$ and $P_{T\rightarrow S} = F(I_{T\rightarrow S})$. More specifically, we define the cross domain consistency loss as 

\begin{equation}
\label{crdoco loss}
\begin{split}
 L&_{crdoco}(G_{T\rightarrow S}, F, X_T) \\
 &= \mathbb{E}_{I_T \sim X_T}[\|F(I_T) - F(G_{T\rightarrow S}(I_T))\|_1].
\end{split} 
\end{equation}

%-------------------------------------------------------------------------
\paragraph{Full objective.}
The overall objective function in the style transfer stage is defined as

\begin{equation}
\label{loss:stage1}
\begin{split}
 L_{style\_transfer} &= L_{adv} + \lambda_{cycle} \cdot L_{cycle} \\
                     &+ \lambda_{identity} \cdot L_{identity} \\
                     &+ \lambda_{task} \cdot L_{task} + \lambda_{smooth} \cdot L_{smooth} \\
                     &+ \lambda_{crdoco} \cdot L_{crdoco},
\end{split} 
\end{equation}

where each $\lambda$ controls the relative importance of each objective. 
Then, we optimize the following min-max problem in the style transfer stage:

\begin{equation}
\label{min-max game}
\begin{split}
 F^* = \arg\min_{F} \min_{\substack{G_{S\rightarrow T} \\ G_{T\rightarrow S}}} \max_{\substack{D_S, \\ D_T}} L_{style\_transfer}.
\end{split} 
\end{equation}

In later stages, we leverage $G_{S\rightarrow T}$ and $G_{T\rightarrow S}$ pre-trained in this style transfer stage.

\subsection{Depth estimation stage}
As the last stage in our UDA framework, we fine-tune the image-to-depth task network $F$ trained in the SSRL stage, by using both synthetic and real-world datasets as shown in Fig 2 in our main paper. The network architecture in this stage is similar to that in the style transfer stage but the weights of the image-to-image translation networks $G_{S\rightarrow T}$ and $G_{T\rightarrow S}$ are fixed for faster training. Specifically, We train $F$, \ie both $F_{enc}$ and $F_{dec}$, by minimizing the following objectives:

\begin{equation}
\label{loss:depth estimation}
\begin{split}
 L_{depth\_estimation} &= \lambda_{task} \cdot L_{task} + \lambda_{smooth} \cdot L_{smooth} \\
                       &+ \lambda_{crdoco} \cdot L_{crdoco},
\end{split} 
\end{equation}

where each $\lambda$ controls the relative importance of each objective.

%-------------------------------------------------------------------------

\section{Training detail}
In this section, we provide more details of our hyper-parameter setting and training strategy.

\subsection{Hyper-parameters}
As mentioned in Section 5 in our main paper, we set the relative weights of the different loss functions based on previous works and our experiments. Specifically, we follow \cite{CycleGAN2017} to set $\lambda_{cycle} = 10$ for our bidirectional image-to-image translation network. Also, similar to~\cite{zheng2018t2net}, we set $\lambda_{identity}=100$, $\lambda_{task}=100$, $\lambda_{smooth}=0.1$. Lastly, we set $\lambda_{crdoco}=1$ based on our ablation study as in Table~\ref{ablation study crdoco}.

%-------------------------------------------------------------------------

\begin{table}[t]
\begin{center}
\scalebox{1.0}{
\begin{tabular}{ccc}
\hline
Stage   & Output channels     & Learning rate        \\ \hline 
1       & 48           & lr / $10^3$ \\ 
2       & 24           & lr / $10^3$ \\ 
3       & 40           & lr / $10^3$ \\ 
4       & 64           & lr / $10^3$ \\ \hdashline
5       & 128          & lr / $10^2$ \\ 
6       & 176          & lr / $10^2$ \\ \hdashline
7       & 304          & lr / $10^1$ \\ 
8       & 512          & lr / $10^1$ \\ 
9       & 2048         & lr / $10^1$ \\ \hline
\end{tabular}
}
\caption{Details of differential learning rates applied to EfficientNet-B5~\cite{tan2019efficientnet} pre-trained on ImageNet~\cite{imagenet_cvpr09} as the encoder $F_{enc}$ of the task network $F$. Note that we set a base learning rate $lr=0.0004$ for $F$ as mentioned in Section 5 in our main paper.}
\label{dflr}
\end{center}
\end{table}

\subsection{Encoder with ImageNet initialization}
As mentioned in Section 5 in our main paper, we leverage EfficientNet-B5~\cite{tan2019efficientnet} pre-trained on ImageNet~\cite{imagenet_cvpr09} as the encoder $F_{enc}$ of the task network $F$.
EfficientNet-B5 mainly consists of 9 stages and each stage yields feature maps with a different number of channel and resolution sizes. To build the encoder-decoder architecture, we remove its last dense layer. Note that we follow previous works~\cite{bhat2020adabins, lee2019big, zheng2018t2net, CrDoCo} to utilize skip connections~\cite{ronneberger2015u}. 

During optimization, we utilize differential learning rates for the encoder $F_{enc}$ based on the stage as shown in Table~\ref{dflr}. 
More specifically. we use relatively lower learning rates for the initial few stages since these stages are already good at extracting general information, such as edges, through ImageNet initialization. By contrast, the last few stages are trained with relatively higher learning rates to enable $F_{enc}$ to adopt to our depth estimation task. Note that we train our comparison methods using our task network $F$ together with the ImageNet initialization and the differential learning rates for a fair comparison. Please refer to our implementation code for more details.

\section{Additional qualitative result}
We provide additional qualitative results in Fig~\ref{fig:additional qualitative results}, highlighting relatively successful and failure cases. From the successful cases, our method is better at estimating consistent depth values on objects' surfaces than comparable methods~\cite{zheng2018t2net, CrDoCo}. It is also worth analyzing the failure cases for future research. As a common trend, current UDA methods including our method fail to handle reflective surfaces (or overexposed white regions) in images. We suspect that this is because the larger surfaces with such a uniform color do not provide useful information for depth estimation. One possible solution would be an introduction of an attention mechanism to utilize global information between pixels.

% \section{Broader impact}
% \label{broader impact}
% Our work proposes a general method for depth estimation and a general component 
% for self-supervised representation learning. We do not work on a specific application, so we do not think there are any foreseeable societal consequences specific to our method.

% {\small
% \bibliographystyle{ieee_fullname}
% \bibliography{wacv}
% }

\end{document}